%% file: iclr2025_scifm_workshop.tex
\documentclass{article}
\usepackage{iclr2025_scifm_workshop,times}
\usepackage{booktabs}
\usepackage[caption=false]{subfig}
\usepackage{graphicx}

\input{math_commands.tex}

\usepackage{hyperref}
\usepackage{url}
\usepackage{multicol}
\usepackage{pifont}

\newcommand{\makecell}[2][@{}c@{}]{\begin{tabular}{#1}#2\end{tabular}}
\newcommand{\nce}{\mathrm{nce}}

\newcommand{\jembedIII}{\href{https://huggingface.co/jinaai/jina-embeddings-v3}{\texttt{jina-embeddings-v3}}}
\newcommand{\jclipI}{\href{https://huggingface.co/jinaai/jina-clip-v1}{\texttt{jina-clip-v1}}}
\newcommand{\jclipII}{\href{https://huggingface.co/jinaai/jina-clip-v2}{\texttt{jina-clip-v2}}}

\title{\jclipII: Multilingual Multimodal \\ Embeddings for Text and Images}

\author{
\begin{tabular}{@{}l}
Andreas Koukounas\thanks{Equal contribution}\quad Georgios Mastrapas$^{*}$\quad Sedigheh Eslami\quad Bo Wang\quad \\ Mohammad Kalim Akram\quad Michael G\"unther\quad Isabelle Mohr\quad Saba Sturua\quad \\ Nan Wang\quad Han Xiao \\
\end{tabular}\\
\\
Jina AI GmbH,  \\
Prinzessinnenstr. 19-20, 10969\\
Berlin, Germany \\
\texttt{research@jina.ai}
}

\iclrfinalcopy % Uncomment for camera-ready version, but NOT for submission.

\begin{document}

\maketitle

\begin{abstract}

Contrastive Language-Image Pretraining (CLIP) has been widely used for crossmodal information retrieval and multimodal understanding tasks.
However, CLIP models are mainly optimized for crossmodal vision-language tasks and underperform in single-mode text tasks. Moreover, these models are often trained on English datasets and therefore lack multilingual understanding. Additionally, from a visual understanding perspective, previous CLIP-based models exhibit insufficient understanding of visually rich documents.
In this work, we propose \jclipII{}, a contrastive vision-language model trained on text pairs, triplets and image-text pairs via a multi-task and multi-stage contrastive learning paradigm in order to support both text-only and crossmodal tasks. We employ a multilingual text encoder and expand the training dataset to include multilingual texts from 29 non-English languages, including Hindi, Chinese, German, French, and others, as well as images of visually-rich documents. 
We evaluate the model’s performance and show that \jclipII{} achieves notable improvements over state-of-the-art CLIP-based models in zero-shot text-only retrieval, semantic textual similarity, and crossmodal retrieval tasks in both English and multilingual settings.
\jclipII{} also provides for flexibility in embedding dimensionality, enabling users to select the granularity of the representations. \jclipII{} is publicly available at \url{https://huggingface.co/jinaai/jina-clip-v2}.

\end{abstract}

\section{Introduction}
\label{sec:introduction}

Contrastive text-image pretraining is a well-known architecture and training framework for building robust text-image alignment models \citep{clip, siglip, evaclip}.
These models are particularly effective for tasks like crossmodal retrieval and zero-shot classification, demonstrating strong generalization capabilities.
However, although CLIP models produce general-purpose embeddings for both texts and images, they are woefully inadequate as a source of text embeddings \citep{jclip1}. We believe this is due to the training regimen of CLIP models, which are typically trained with short and information-poor image captions \citep{longclip} and omit standard techniques for finetuning meaningful text embeddings, e.g. training using hard negatives \citep{entretrieval, rocketqa}.

To overcome these limitations, \citet{jclip1} introduced a multi-task, multi-stage contrastive learning approach to simultaneously align text-text and text-image embeddings.
This method involved three stages: optimizing text-text and text-image embeddings for short image-caption and text-text pairs (stage 1), refining alignment using long text-text pairs and detailed image-caption pairs (stage 2), and further enhancing performance with text triplets containing hard negatives and detailed image-caption pairs (stage 3). The resulting model, \jclipI{}, achieves strong performance on both the crossmodal CLIP Benchmark\footnote{https://github.com/LAION-AI/CLIP\_benchmark} and the text embedding MTEB Benchmark \citep{mteb}. Despite its strengths, \jclipI{} \citep{jclip1} has several limitations.
First, it is an English-only model, which makes it unsuitable for multilingual document retrieval.
Second, \jclipI{} struggles with visually-rich images, e.g. those containing text, tables, graphs, and diagrams \citep{colpali}. Moreover, \jclipI{} only supports fixed length embeddings, which can be wasteful when embedding vectors are much larger than required for a task \citep{mrl}.

To address these challenges, we propose several enhancements to the pretraining scheme.
For multilingual support, our approach uses a multilingual language model \citep{xlmroberta} to initialize the text encoder, and incorporates multilingual text-image and text-text pairs during training. To improve performance on visually complex documents, we gradually increase image resolution during pretraining and use multimodal training pairs with complex visual structures. Finally, to enable users to select the granularity of their representations, we employ \textit{Matryoshka Representation Learning} \citep{mrl}
and enable dimensional truncation on the output vectors with minimal performance degradation.

\begin{figure*}[t]
\centering
\includegraphics[width=\linewidth]{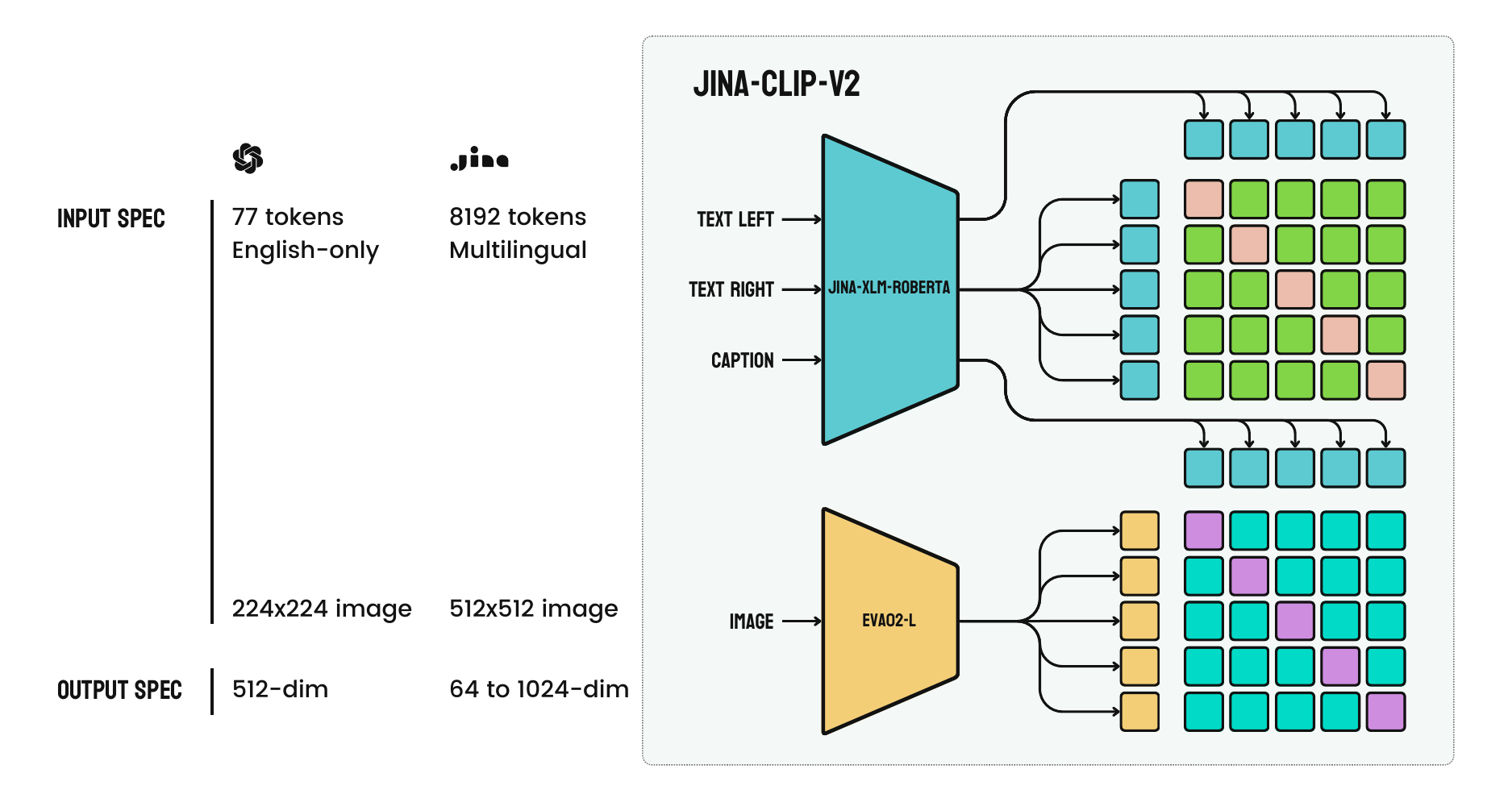}
\caption{\jclipII{} combines a text encoder (Jina XLM-RoBERTa, 561M parameters) and a vision encoder (EVA02-L14, 304M parameters) for a total of 865M parameters.}
\label{img:clip-arch}
\end{figure*}

The resulting model, \jclipII{} depicted in Figure~\ref{img:clip-arch}, not only achieves competitive performance on crossmodal retrieval benchmarks against multilingual state-of-the-art models like NLLB-CLIP \citep{nllbclip} but also performs comparably to dedicated multilingual text embedding models like \jembedIII{} \citep{jembeddings3} on the retrieval and semantic textual similarity (STS) tasks of the MTEB benchmark \citep{mteb}. Moreover, due to the inclusion of visually rich training data and the progressive increase in image resolution, \jclipII{} demonstrates significantly improved performance on ViDoRe, a benchmark for visually rich document retrieval \citep{colpali}, compared to \jclipI{}. In summary, our contributions are the following:

\begin{enumerate}
    \item \textbf{Support for multiple languages}. \jclipII{} achieves up to 67\% higher scores on multilingual crossmodal performance, up to 60\% on multilingual Retrieval performance and up to 43\% on multilingual STS, compared \jclipI{}.
    \item \textbf{Improved performance on visual document retrieval}. \jclipII{}'s scores are 35\% higher on the ViDoRe benchmark \citep{colpali} compared to \jclipI{}.
    \item \textbf{Flexible embedding dimensionality}. Truncating \jclipII{} embeddings by up to 75\% (from 1,024 dimensions down to 256)  reduces performance by \textless1\%.
\end{enumerate}

\section{Related Work}
\label{sec:related-work}
%\vspace{-5pt}

Previous work \citep{align, albef, clip, vl_contrastive_medical} pioneered dual-encoder architectures trained contrastively on image-text pairs, demonstrating impressive zero-shot performance and laying the groundwork for pretrained vision-language models. Building on this foundation, \citet{siglip} proposed an alternative and more efficient sigmoid objective, while \cite{evaclip} optimized the model's dimensions and scaled up both dataset and model sizes, achieving state-of-the-art performance on crossmodal tasks. \citet{jclip1} enhanced a CLIP-based retriever with strong text-to-text and crossmodal retrieval capabilities, however, their work fails to address retrieval scenarios involving candidate pools with heterogeneous modalities, due to the modality gap \citep{modgap, alignclip} and the modality bias \citep{mmembed}. To mitigate the modality bias, \cite{mmembed} finetuned a multimodal LLM with modality-aware hard-negative mining, creating a universal multimodal retriever and maintaining competitive text-to-text retrieval performance across diverse tasks. In this work, we build on the training strategy proposed by \cite{jclip1} and train our model using a text-to-text as well as a text-to-image contrastive InfoNCE \citep{infonce} loss.

Extensive research has also focused on extending CLIP to languages beyond English.
To address the lack of image-caption pairs in languages other than English, \citet{mclip} applied knowledge distillation to retrain the text encoder using machine-translated data.
NLLB-CLIP \citep{nllb-clip} leverages \textit{Locked-image Tuning (LiT)} \citep{lit} alongside the multilingual NLLB \citep{nllb} text encoder, achieving state-of-the-art results in both retrieval and classification tasks.

Multilingual text embedding models are typically based on either an encoder-only architecture, such as XLM-RoBERTa \citep{xlmroberta}, or a decoder-only multilingual large language model, such as Mistral 7B \citep{mistral7b}.
Multilingual E5 \citep{me5} and BGE-M3 \citep{bgem3} both use XLM-RoBERTa as their backbone, the first leveraging extensive multilingual training with instruction tuning, and the second employing multi-task learning techniques.
mGTE \citep{mgte} achieves comparable performance to BGE-M3 using a smaller transformer architecture, enhanced by RoPE \citep{rope} to extend the input context length.
Similarly, \citet{jembeddings3} employ RoPE to increase the model's context length and apply LoRA tuning \citep{lora} to optimize embeddings for downstream tasks.

Finally, \citet{mrl} propose a training technique, which they call \textit{Matryoshka Representation Learning}, that enables embedding models to learn coarse-to-fine representations. These representations can be truncated during inference to match the requirements of downstream tasks, cutting down on potentially redundant vector dimensions.

\section{\jclipII}
\label{sec:training}

The \jclipII{} model uses the dual encoder architecture, introduced in the original CLIP \citep{clip} model and reused in \jclipI{} \citep{jclip1}.

The text encoder is initialized with pretrained Jina-XLM-RoBERTa model weights. Introduced in \jembedIII{} \citep{jembeddings3}, the Jina-XLM-RoBERTa model is a port of the multilingual XLM-RoBERTa \citep{xlmroberta} model to a modern encoder-only architecture with Flash Attention \citep{flashattention}, rotary positional embeddings \citep{rope} and support for low-rank adaptation \citep{lora}.

Like \jclipI{}, the image encoder is a pretrained EVA02 model \citep{eva02}. We selected the L/14 pretrained model variant, which is similar in number of parameters to the text encoder. This model implementation includes 2D rotary positional embeddings and a memory-efficient attention implementation based on \textit{xFormers} \citep{xformers}. 
Table \ref{tab:model-properties} presents architectural details for both encoders.

\begin{table}[t]
    \centering
    \caption{Model properties} 
    \label{tab:model-properties}
    \setlength{\tabcolsep}{4.5pt} 
\vskip 0.01in
\small{
\begin{tabular}{ l|c|c }
 \toprule
 \textbf{Feature} & \textbf{Text Encoder} & \textbf{Image Encoder} \\
 \midrule
Base Model & Jina-XLM-RoBERTa \citep{jembeddings3} & EVA02 L/14 \citep{evaclip} \\
Parameters & 561M & 304M \\
Input Specification	& 8,192 tokens (max) & $\mathrm{(512, 512)}$ resolution \\
Output Dimensions & 1,024 & 1,024 \\
Layers & 24 & 24 \\
Attention Implementation & FlashAttention2 \citep{flashattention} & xFormers \citep{xformers} \\
Pooling Strategy & Mean pooling	& CLS pooling \\
Additional Features	& 89 languages supported & Patch size 14x14 \\
 \bottomrule
\end{tabular}
}
\end{table}

\subsection{Training Data}
\label{sec:datasets}

Similarly to \jclipI{} \citep{jclip1}, we constructed four training datasets used in different stages of the training process. Dataset $\mathbb{D}^\mathrm{txt;p}$ denotes a dataset of text pairs, $\mathbb{D}^\mathrm{txt;t}$ a dataset of text samples with hard negatives, $\mathbb{D}^\mathrm{mm;s}$ a multimodal short-caption dataset and $\mathbb{D}^\mathrm{mm;l}$ a multimodal long-caption dataset.

Both $\mathbb{D}^\mathrm{txt;p}$ and $\mathbb{D}^\mathrm{txt;t}$ contain data in 30 languages drawn from various existing datasets for text information retrieval and semantic similarity, and were introduced as training data for \jembedIII{} \citep{jembeddings3}. $\mathbb{D}^\mathrm{txt;p}$ consists of pairs of texts drawn from a diverse collection of datasets, and $\mathbb{D}^\mathrm{txt;t}$ is a high-quality dataset with hard negatives from various sources, where each sample contains one annotated positive and seven negative items. \citet{jembeddings3} provide detailed information about these datasets and how they were curated. 

Our multimodal training datasets draw on multiple sources. We randomly sampled $\sim$400M image-caption pairs from the $\sim$2B in the DFN dataset \citep{dfn} to obtain a collections of image-text pairs in English. To this, we added non-English image-text pairs sampled from CommonPool \citep{datacomp}. We filtered this data by language, aspect ratio, and by removing any caption with less than five words. We used multilingual SigLIP \citep{siglip} to rank image-text pairs based on cosine similarity and kept the top scoring $\sim$50\%. The result was an additional $\sim$400M image-text pairs. All images were resized to $\mathrm{(384, 384)}$. These two large-scale datasets are included in $\mathbb{D}^\mathrm{mm;s}$ and a small part with captions longer than 256 tokens is held out for $\mathbb{D}^\mathrm{mm;l}$.

In order to overcome the limitations of visual document understanding, we diversify our training data with PDFs, scientific graphs, infographics and Wikipedia images. Specifically, the short caption image-text dataset $\mathbb{D}^\mathrm{mm;s}$ and the long-caption $\mathbb{D}^\mathrm{mm;l}$ include the following training datasets: DocVQA \citep{docvqa}, TatDQA \citep{tatdqa}, InfographicsVQA \citep{infographicvqa}, SciGraphQA \citep{scigraphqa}, ArXivQA and ArXivCAP \citep{mmarxiv}, WIT \citep{wit} and ViDoRe synthetic training data \citep{colpali}. For the QA datasets, we concatenated the query and answer for each sample to construct a matching text in lieu of a caption. For the WIT dataset, we use the reference caption as corresponding text in $\mathbb{D}^\mathrm{mm;s}$ and a concatenation of reference caption, page title, section title, page description and section description as matching text for $\mathbb{D}^\mathrm{mm;l}$.

Finally, $\mathbb{D}^\mathrm{mm;l}$ includes multilingual synthetic long captions, similar to how \citet{jclip1} make use of ShareGPT4v \citep{sharegpt4v}. We use the GPT4v API \citep{gpt4v} to generate detailed image descriptions for 40,000 images in 30 languages. This adds $\sim$1.2M generated multilingual long captions to $\mathbb{D}^\mathrm{mm;l}$.

\subsection{Training Stages}
\label{sec:training-stages}

Inspired by \citet{jclip1}, we employ a multi-task, multi-stage training approach, in order to optimize the model for two tasks simultaneously: text-image matching and text-text matching.

\textbf{Stage 1} focuses on aligning the multimodal representations while also improving the text representations of the text encoder. We train on $\mathbb{D}^\mathrm{txt;p}$ and $\mathbb{D}^\mathrm{mm;s}$, with a small context length of 77 and an image resolution of $\mathrm{(224, 224)}$ to enable large batch sizes of $16,384$. This is the most computationally expensive stage in our training pipeline, requiring approximately 10 days on 8 NVIDIA H100 GPUs to converge.  Towards the end, when performance peaks, we switch to a higher image resolution of $\mathrm{(384, 384)}$ using positional embedding interpolation as a warm-up for the next stage.

\textbf{Stage 2} trains on $\mathbb{D}^\mathrm{txt;p}$ and $\mathbb{D}^\mathrm{mm;l}$, with context length increased from 77 to 512 and image resolution set at $\mathrm{(384, 384)}$. This stage improves text embedding performance on longer text lengths, while maintaining alignment between the two modalities.

\textbf{Stage 3} uses hard negatives from $\mathbb{D}^\mathrm{txt;t}$ to further improve the text encoder in distinguishing relevant from irrelevant text. To maintain text-image alignment, we continue training with $\mathbb{D}^\mathrm{mm;l}$. We do one more positional embedding interpolation to increase the image resolution to $\mathrm{(512, 512)}$ and train on this resolution for the duration of stage 3.

Table \ref{appendix:training-settings} in Section~\ref{appendix} specifies the training details for each step.

\subsection{Loss Functions}
\label{sec:loss-functions}

Given a batch $\mathbf{B} \subset \mathbb{D}^\mathrm{pairs}$ of embedding pairs $(\bf{q},\bf{p})$, the InfoNCE loss $\mathcal{L}_{\nce}$ \citep{infonce}, given in \eqref{eq:loss-pairs}, evaluates the cosine similarity $cos(\bf{q},\bf{p})$ between query $\bf{q} \in \mathbb{R}^{d}$ and its corresponding target $\bf{p} \in \mathbb{R}^{d}$, relative to the similarity of all other targets in the batch. The loss is calculated in both directions to preserve the symmetry of similarity measures. The temperature parameter $\tau$ influences how the loss function weighs minor differences in the similarity scores.
%
%\begin{flalign}
%    & \mathcal{L}_{\nce}(\mathbf{B}) := \mathcal{L}_{\nce}^{\longrightarrow}(\mathbf{B}) + \mathcal{L}_{\nce}^{\longleftarrow}(\mathbf{B}),\text{ with} \nonumber \\
%    & \mathcal{L}_{\nce}^{\longrightarrow}(\mathbf{B}) := \mathbb{E}_{(\bf{q},\bf{p})\sim \mathbf{B}}\left[-\ln \frac{e^{cos(\bf{q}, \bf{p})/\tau}}{\sum\limits_{i = 1}^k e^{cos(\bf{q}, \bf{p_i})/ \tau}}\right] \nonumber \\
%    & \mathcal{L}_{\nce}^{\longleftarrow}(\mathbf{B}) := \mathbb{E}_{(\bf{q},p)\sim \mathbf{B}}\left[-\ln \frac{e^{cos(\bf{p}, \bf{q}) / \tau}}{\sum\limits_{i = 1}^k e^{cos(\bf{p}, \bf{q_i}) / \tau}}\right]
%\label{eq:loss-pairs}
%\end{flalign}
%
\vspace{+2mm}
\begin{gather}
\mathcal{L}_{\nce}(\mathbf{B}) := 
 \mathcal{L}_{\nce}^{\longrightarrow}(\mathbf{B}) + \mathcal{L}_{\nce}^{\longleftarrow}(\mathbf{B}) \quad \text{where,} \notag \\
 \mathcal{L}_{\nce}^{\longrightarrow}(\mathbf{B}) := \mathbb{E}_{(\bf{q},\bf{p})\sim \mathbf{B}}\left[-\ln \frac{e^{cos(\bf{q}, \bf{p})/\tau}}{\sum\limits_{i = 1}^k e^{cos(\bf{q}, \bf{p_i})/ \tau}}\right] , 
 \mathcal{L}_{\nce}^{\longleftarrow}(\mathbf{B}) := \mathbb{E}_{(\bf{q},p)\sim \mathbf{B}}\left[-\ln \frac{e^{cos(\bf{p}, \bf{q}) / \tau}}{\sum\limits_{i = 1}^k e^{cos(\bf{p}, \bf{q_i}) / \tau}}\right]
\label{eq:loss-pairs}
\end{gather}
When hard negatives are available for query $q$, an extended version of the $\mathcal{L}_{\nce}$ loss is used. Given a batch $\mathbf{B} \subset \mathbb{D}^\mathrm{triplets}$ of samples $r = (\bf{q}, \bf{p}, \bf{n_1} ..., \bf{n_7})$, consisting of a query $\bf{q}$, a positive match $\bf{p}$, and seven negatives $\bf{n_1} ..., \bf{n_7}$, the extended loss function, denoted here as $\mathcal{L}_{\nce^+}$, is given in \eqref{eq:loss-hard-negatives}. Similarly to $\mathcal{L}_{\nce}$, this loss function is bidirectional.
%
%\begin{flalign}
%&\mathcal{L}_{\nce^+}(\mathbf{B}) := \nonumber\\
%&\;\;\;\;\;\mathbb{E}_{r\sim \mathbf{B}}\Bigg[-\ln \frac{e^{cos(\bf{q}, \bf{p}) / \tau}}{\sum\limits_{i = 1}^k \Big[ e^{cos(\bf{q}, p_i) / \tau}+ \sum\limits_{j = 1}^{7} e^{cos(\bf{q}, \bf{n_{j,i}}) / \tau}\Big]}\Bigg]\nonumber \\
%&\, + \mathbb{E}_{r\sim \mathbf{B}}\Bigg[-\ln \frac{e^{cos(\bf{p}, \bf{q}) / \tau}}{\sum\limits_{i = 1}^k e^{cos(\bf{p}, \bf{q_i}) / \tau}}\Bigg]\nonumber \\
%&\text{with}\; r = (\bf{q},\bf{p}, \bf{n_1}, \ldots, \bf{n_{7}}).\label{eq:loss-hard-negatives}
%\end{flalign}
%
\vspace{+2mm}
\begin{gather}
\mathcal{L}_{\nce^+}(\mathbf{B}) :=
\mathbb{E}_{r\sim \mathbf{B}}\Bigg[-\ln \frac{e^{cos(\bf{q}, \bf{p}) / \tau}}{\sum\limits_{i = 1}^k \Big[ e^{cos(\bf{q}, p_i) / \tau}+ \sum\limits_{j = 1}^{7} e^{cos(\bf{q}, \bf{n_{j,i}}) / \tau}\Big]}\Bigg] + %\notag \\
\mathbb{E}_{r\sim \mathbf{B}}\Bigg[-\ln \frac{e^{cos(\bf{p}, \bf{q}) / \tau}}{\sum\limits_{i = 1}^k e^{cos(\bf{p}, \bf{q_i}) / \tau}}\Bigg]
\label{eq:loss-hard-negatives}
\end{gather}
Equation~\ref{eq:loss-stages} defines the loss for each of the three stages. In each stage, we optimize a joint loss function, i.e. the sum of two loss functions, one operating on text representations and one on multimodal representations. In equation~\ref{eq:loss-stages}, $\mathbf{B_\mathit{k}}$ denotes a batch of samples drawn from the dataset $\mathbb{D}^\mathrm{k}$, where $\mathrm{k} \in \{\mathit{txt;p}, \mathit{txt;t}, \mathit{mm;s}, \mathit{mm;l}\}$. All stages optimize $\mathcal{L}_{\nce}$ except Stage 3, which uses hard negatives in the text loss branch and thus calculates $\mathcal{L}_{\nce^+}$. 
\vspace{+1mm}
\begin{flalign}
&\mathcal{L}_1(\mathbf{B_\mathit{txt;p}}, \mathbf{B_\mathit{mm;s}}) := \mathcal{L}_{\nce}(\mathbf{B_\mathit{txt;p}}) + \mathcal{L}_{\nce}(\mathbf{B_\mathit{mm;s}}) \nonumber\\
&\mathcal{L}_2(\mathbf{B_\mathit{txt;p}}, \mathbf{B_\mathit{mm;l}}) := \mathcal{L}_{\nce}(\mathbf{B_\mathit{txt;p}}) + \mathcal{L}_{\nce}(\mathbf{B_\mathit{mm;l}}) \nonumber\\
&\mathcal{L}_3(\mathbf{B_\mathit{txt;t}}, \mathbf{B_\mathit{mm;l}}) := \mathcal{L}_{\nce^+}(\mathbf{B_\mathit{txt;t}}) + \mathcal{L}_{\nce}(\mathbf{B_\mathit{mm;l}})
\label{eq:loss-stages}
\end{flalign}

\subsection{Matryoshka Representation Learning}
\label{subsec:matryoshka-representation-learning}

In every training stage, the loss component is recalculated at different dimensionalities: Loss is computed for the full 1024-dim output vector, and for truncated subsets of 64, 128, 256, 512 and 768 dimensions. Model weight adjustment proceeds on the basis of all those losses combined.
This training technique, called \textit{Matryoshka Representation Learning} \citep{mrl}, induces the model to encode its embeddings by increasing granularity.
This makes it possible, at inference time, to truncate embedding vectors with a minor performance penalty.

\section{Evaluation}
\label{sec:evaluation}

We evaluate the performance of \jclipII{} on a selection of benchmarks: English and multilingual crossmodal retrieval benchmarks in Section~\ref{sec:crossmodal-evals}, text embedding tasks from the MTEB benchmark suite in Section~\ref{sec:mteb-evals} and visual document retrieval tasks in Section~\ref{sec:vidore-evals}. Finally, we present an ablation study of \jclipII{}'s Matryoshka representations in Section~\ref{sec:mrl-evals}. Detailed evaluation results are presented in Section~\ref{appendix} tables \ref{appendix:clipb} to \ref{appendix:mrl-mteb}.

\subsection{Crossmodal Retrieval Evaluation}
\label{sec:crossmodal-evals}

\begin{table*}[t]
    \centering
    \caption{Evaluation results on crossmodal retrieval tasks} 
    \label{tab:results-cm}
    \setlength{\tabcolsep}{4.5pt} 
\vskip 0.01in
\small{
\begin{tabular}{ lc|cc|cc|cc}
 \toprule
 \multicolumn{2}{l|}{\textbf{Benchmark}} &  \multicolumn{2}{c|}{\textbf{CLIP Benchmark}} & \multicolumn{2}{c|}{\textbf{Crossmodal-3600}} & \multicolumn{2}{c}{\textbf{XTD10}} \\
 \midrule
 \multicolumn{2}{l|}{\textbf{Language}} & \multicolumn{2}{c}{English} & \multicolumn{2}{c}{Multilingual} & \multicolumn{2}{c}{Multilingual} \\
 \midrule
 \multicolumn{2}{l|}{\textbf{Task Type}} & \multicolumn{6}{c}{Zero-Shot Retrieval} \\
 \midrule
 \textbf{Model} & \textbf{\#Parameters} & T-I r@5 & I-T r@5 &T-I r@5 & I-T r@5  & T-I r@5 & I-T r@5 \\
 \midrule 
 \jclipI & 223M & 77.75 & 87.65 & 16.93 & 19.82 & 31.01 & 36.89 \\
 nllb-siglip-base & 507M & 79.57 & 86.38 & 79.29 & 76.56 & 86.23 & 84.87\\
 nllb-siglip-large & 1.2B & \textbf{81.54} & 88.15 & 82.07 & 80.16 & \textbf{87.60} & 85.37 \\
 \midrule 
 \jclipII stage 1 & 865M & 73.87 & 86.61 & 73.51 & 79.64 & 80.66 & 83.02 \\
 \jclipII stage 2 & 865M & 79.81 & 89.57 & \textbf{84.13} & \textbf{84.12} & 86.11 & \textbf{86.45}\\
 \jclipII & 865M & 79.09 & \textbf{89.73}  & 81.43 & 83.23 & 84.87  & 86.03 \\
 \bottomrule
\end{tabular}
\vspace{0.01in}
\\ T-I r@5: Text to Image Recall@5 [\%]\quad{}
I-T r@5: Image to Text Recall@5 [\%]\quad{} 
}
\end{table*}

We evaluate \jclipII{} on a set of English and multilingual crossmodal tasks, conducting a comparative analysis with both \jclipI{} and the state-of-the-art NLLB-CLIP \citep{nllbclip} large and base SigLIP variants.
Additionally, we track the model's performance at all training stages to evaluate the effectiveness of the training protocol.
For English zero-shot image-text and text-image retrieval, we conduct evaluations on the CLIP Benchmark, which includes Flickr30K~\citep{flickr30k} and COCO Captions~\citep{mscococaptions}. For multilingual crossmodal retrieval tasks, we assess performance on the Crossmodal-3600 \citep{crossmodal3600} and XTD10 \citep{xtd10, mic} datasets.

Table \ref{tab:results-cm} demonstrates the strong performance of \jclipII{} across both English and multilingual benchmarks. On the English CLIP Benchmark, our model outperforms \jclipI{} and while \texttt{nllb-siglip-large} leads in text-to-image retrieval, \jclipII{} outperforms both NLLB-CLIP variants in image-to-text retrieval. On multilingual multimodal retrieval, we obtain competitive results, approaching the performance of \texttt{nllb-siglip-large} in text-to-image retrieval and surpassing it in image-to-text retrieval. Regarding stage progression, performance improves considerably between Stage 1 and Stage 2 on all benchmarks, but drops slightly from Stage 2 to Stage 3. We hypothesize that there is a trade-off between crossmodal and text-only retrieval performance, and during stage 3 training, where focus shifts towards text embedding training with hard negatives, the model falls slightly behind on crossmodal retrieval.

\subsection{Text Retrieval and STS Evaluation}
\label{sec:mteb-evals}

% \begin{table*}[ht]
%     \centering
%     \caption{Evaluation results on MTEB \citep{mteb} Retrieval and STS tasks}
%     \label{tab:results-mteb}
%     \setlength{\tabcolsep}{4.5pt} 
% \vskip 0.01in
% \small{
% \begin{tabular}{lcc|cc|cc}
%  \toprule
%  \multicolumn{3}{l|}{\textbf{Task Type}} & \multicolumn{2}{c|}{\textbf{Retrieval}}   & \multicolumn{2}{c}{\textbf{STS}} \\
%  \midrule
%  \multicolumn{3}{l|}{\textbf{Metric}} & \multicolumn{2}{c|}{nDCG@10 [\%]}   & \multicolumn{2}{c}{Spearman$^*$ [\%]} \\
%  \midrule
%  \textbf{Model} & \textbf{\#Parameters$^{**}$} & \textbf{Multimodal} & English & Multilingual & English & Multilingual \\
%  \midrule
%  \jclipI & 137M & \ding{51} & 48.33 & 42.81 & 80.92 & 55.71 \\
%  \texttt{nllb-siglip-large} & 767M & \ding{51} & 24.91 & 39.95 & 74.89 & 63.63  \\
%  \jembedIII & 572M & \ding{55} & \textbf{53.87} & \textbf{72.58} & \textbf{85.80} & 69.81 \\
%  \midrule
%  \jclipII stage 1 & 561M & \ding{51} & 40.91 & 53.81 & 79.67 & 71.42 \\
%  \jclipII stage 2 & 561M & \ding{51} & 43.17 & 56.67 & 80.33 & 71.68 \\
%  \jclipII & 561M & \ding{51} & 49.32 & 58.90  & 81.29 & \textbf{73.29} \\
%  \bottomrule
% \end{tabular}
% \\
% \footnotesize{$^*$ Spearman correlation based on cosine similarity \quad{} $^{**}$ Refers to the parameters of the text tower}
% \\
% }
% \end{table*}

\begin{table*}[ht]
  \centering
  \caption{Evaluation results on MTEB \citep{mteb} Retrieval and STS tasks across 8 languages}
  \label{tab:results-mteb}
  \setlength{\tabcolsep}{4.5pt}
  \vskip 0.1in
  \small{
    \begin{tabular}{lc|cccccccc}
    \toprule
    \multicolumn{10}{c}{\textbf{Retrieval} nDCG@10 [\%]} \\
    \midrule
    \textbf{Model} & \textbf{\#Parameters$^*$} & en & zh & hi & de & fr & es & jp & ru \\
    \midrule
    \jclipI & 137M & 47.04 & 8.09 & 8.73 & 33.77 & 39.99 & 39.77 & 10.77 & 4.78 \\
    nllb-siglip-base & 414M & 24.69 & 27.04 & 39.18 & 22.67 & 28.30 & 30.06 & 30.73 & 32.68 \\
    nllb-siglip-large & 767M & 27.83 & 32.96 & 43.22 & 32.29 & 38.74 & 39.92 & 35.53 & 38.87 \\
    \jembedIII & 572M & \textbf{48.34} & \textbf{64.27} & \textbf{65.18} & \textbf{63.59} & \textbf{64.09} & \textbf{62.97} & \textbf{68.47} & \textbf{66.46} \\
    \jclipII stage 1 & 561M & 42.02 & 55.11 & 62.03 & 57.56 & 58.56 & 59.86 & 63.34 & 60.74 \\
    \jclipII stage 2 & 561M & 43.47 & 57.66 & 62.19 & 59.16 & 61.05 & 61.51 & 65.13 & 63.66 \\
    \jclipII & 561M & 46.46 & 60.47 & 63.08 & 56.66 & 61.16 & 61.48 & 65.48 & 64.73 \\
    \midrule
    \multicolumn{10}{c}{\textbf{STS} Spearman correlation based on cosine similarity} \\
    \midrule
    \textbf{Model} & \textbf{\#Parameters$^*$} & en & zh & hi & de & fr & es & jp & ru \\
    \midrule
    \jclipI & 137M & 81.35 & 22.97 & 36.09 & 50.04 & 67.97 & 70.05 & 55.83 & 39.77 \\
    nllb-siglip-base & 414M & 72.57 & 33.36 & 68.11 & 39.58 & 63.94 & 65.28 & 77.39 & 57.55 \\
    nllb-siglip-large & 767M & 74.73 & 39.15 & 73.78 & 53.45 & 72.38 & 69.68 & 79.25 & 62.84 \\
    \jembedIII & 572M & \textbf{81.68} & 54.60 & 83.62 & 75.32 & 81.11 & 80.78 & 81.03 & 77.16 \\
    \jclipII stage 1 & 561M & 80.01 & 52.74 & 83.35 & 74.84 & 79.06 & 80.42 & 81.48 & 75.62 \\
    \jclipII stage 2 & 561M & 80.76 & 53.37 & \textbf{84.30} & 75.64 & 80.19 & 80.51 & 81.46 & 76.18 \\
    \jclipII & 561M & 81.58 & \textbf{55.08} & 79.55 & \textbf{76.27} & \textbf{81.70} & \textbf{81.89} & \textbf{82.01} & \textbf{77.79} \\
    \bottomrule
    \end{tabular}
    \\
    \vspace{0.1in}
    \footnotesize{$^*$ Refers to the parameters of the text tower} \\
  }
\end{table*}

Table \ref{tab:results-mteb} presents the results of evaluations on the retrieval and semantic textual similarity (STS) tasks from MTEB \citep{mteb}, both in English-only and multilingual contexts. \jclipII{} significantly improves on the performance of \texttt{nllb-siglip-large} in both English and multilingual text tasks, demonstrating the value of explicit text retrieval training. Compared to \jclipI{}, performance is significantly better on multilingual tasks, and comparable on English-only tasks. Compared to the text-only frontier model \jembedIII{}, performance is lackluster on retrieval tasks. This suggests that multimodal training is hampering text-only performance.

\vspace{-3pt}
\subsection{Visual Document Retrieval}
\label{sec:vidore-evals}
\vspace{-3pt}

\begin{table*}[htb]
    \centering
    \setlength{\tabcolsep}{4pt}
    \caption{ViDoRe benchmark \citep{colpali} evaluation results}
    \label{tab:vidore-benchmark}
    \begin{small}
    \begin{tabular}{l|cccccc}
        \toprule
        Task & \makecell{\texttt{jina-} \\ \texttt{clip-v2}} & \makecell{\texttt{jina-clip-} \\ \texttt{v2 stage 1}} & \makecell{\texttt{jina-clip-} \\ \texttt{v2 stage 2}} & \makecell{\texttt{jina-} \\ \texttt{clip-v1}} & \makecell{\texttt{nllb-siglip} \\ \texttt{large}} & \makecell{\texttt{siglip-} \\ \texttt{so400m}$^{*}$} \\
        \midrule
        \multicolumn{7}{c}{\textbf{Retrieval - nDCG@5 [\%]}} \\
        \midrule
        ArxivQ & \textbf{64.92} & 58.78 & 66.14 & 25.40 & 30.44 & 43.2 \\
        DocQ & 24.64 & 21.16 & 24.81 & 11.90 & 23.82 & \textbf{30.3} \\
        InfoQ & 57.90 & 56.62 & 56.63 & 35.50 & 59.87 & \textbf{64.1} \\
        TabF & 45.91 & 37.28 & 43.18 & 20.20 & \textbf{70.68} & 58.1 \\
        TATQ & \textbf{30.25} & 12.09 & 25.79 & 3.30 & 20.00 & 26.2 \\
        Shift & \textbf{34.07} & 28.41 & 31.54 & 3.80 & 30.79 & 18.7 \\
        AI & \textbf{68.07} & 32.95 & 55.64 & 15.20 & 47.90 & 62.5 \\
        Energy & 62.15 & 49.15 & 60.97 & 19.70 & 64.94 & \textbf{65.7} \\
        Gov & \textbf{68.97} & 37.81 & 55.56 & 21.40 & 58.62 & 66.1 \\
        Health & 69.05 & 39.50 & 57.37 & 20.80 & 58.43 & \textbf{79.1} \\
        \midrule
        \textbf{Average} & \textbf{52.65} & 37.37 & 47.76 & 17.72 & 46.55 & 51.4 \\
        \bottomrule
    \end{tabular}
    \vspace{0.1in}
    \\
    \footnotesize{$^*$ \citep{siglip, sovit}, scores taken directly from \citet{colpali}}
    \\
    \end{small}
\end{table*}
Visual document retrieval challenges vision-language models to capture fine-grained information from images, including embedded text, and accurately match these with text queries or documents. This task is particularly challenging as it deals with high-resolution document images with complex content, which differs significantly from processing typical image-caption datasets. As shown in Table~\ref{tab:vidore-benchmark}, \jclipII{} outperforms other state-of-the-art CLIP models on the ViDoRe Benchmark for visual document understanding \citep{colpali}, with an average nDCG@5 score of 52.65\%, toping \jclipI{}, \texttt{nllb-siglip-large} and \texttt{siglip-so400m-patch14-384} \citep{siglip, sovit}. The progression through training stages brings substantial improvements, highlighting the effectiveness of gradually increasing the image resolution. We further investigate how image resolution affects visual document retrieval in Section \ref{sec:vidore-abl}.

\vspace{-3pt}
\subsection{Matryoshka Representation Learning}
\label{sec:mrl-evals}
\vspace{-3pt}

\begin{table*}[htb]
    \centering
    \caption{MRL \citep{mrl} ablation study on various embedding dimensions}
    \setlength{\tabcolsep}{4.5pt} 
    \label{tab:mrl-ablation}
    \begin{center}
    \begin{small}
    \begin{tabular}{l|cccccc}
    \toprule
   Dataset - Dimension & \makecell{1024} & \makecell{768} & \makecell{512} & \makecell{ 256} & \makecell{128} & \makecell{64} \\
    \midrule
    \multicolumn{7}{c}{\textbf{Text-Image Retrieval - Recall@5  [\%]}} \\
    \midrule
    CLIP Benchmark & 79.10 & \textbf{79.12} & 78.93 & 78.32 & 75.90 & 70.51 \\
    Crossmodal-3600 \citep{crossmodal3600} &  81.43 & \textbf{82.35} & 82.31 & 81.75 & 78.17 & 72.52 \\
    XTD10 \citep{xtd10, mic} & \textbf{84.87} & 84.85 & 84.60 & 84.32 & 81.80 & 77.85 \\
    \midrule
    \multicolumn{7}{c}{\textbf{Image-Text Retrieval - Recall@5  [\%]}} \\
    \midrule
    CLIP Benchmark & \textbf{89.73} & 89.60 & 89.55 & 89.35 & 87.48 & 83.20 \\
    Crossmodal-3600 \citep{crossmodal3600} & 83.23 & \textbf{83.26} & 83.21 & 82.81 & 80.54 & 75.37 \\
    XTD10 \citep{xtd10, mic} & \textbf{86.03} & 86.02 & 86.02 & 85.84 & 84.37 & 81.02 \\
    \midrule
    \multicolumn{7}{c}{\textbf{Text-Text Retrieval - nDCG@10  [\%]}} \\
    \midrule
    EN Classic MTEB \citep{mteb} Retrieval & \textbf{49.33} & 49.32 & 49.19 & 48.67 & 46.37 & 40.66 \\
    \midrule
    \multicolumn{7}{c}{\textbf{Semantic Textual Similarity - Spearman Correlation 
    [\%]}} \\
    \midrule
    EN Classic MTEB \citep{mteb} STS & \textbf{81.29} & 81.27 & 81.26 & 81.24 & 80.78 & 79.56 \\
    \bottomrule
\end{tabular}
\end{small}
\end{center}
\end{table*}

Table \ref{tab:mrl-ablation} presents the impact of embedding truncation on the crossmodal and text-only tasks.
Performance remains highly stable when reducing dimensions from 1024 to 256, with minimal degradation (typically less than 1\%) across all evaluation tasks. At 256 dimensions, representing a 75\% reduction in embedding size, the model preserves effectively all essential semantic information.
Substantial performance degradation is only observed at dimensions 128 and 64.

\section{Analysis}
\label{sec:analysis}

\subsection{The Role of Image Resolution in Visual Document Retrieval}
\label{sec:vidore-abl}
\begin{figure}[h]
    \centering
    \includegraphics[width=0.75\textwidth]{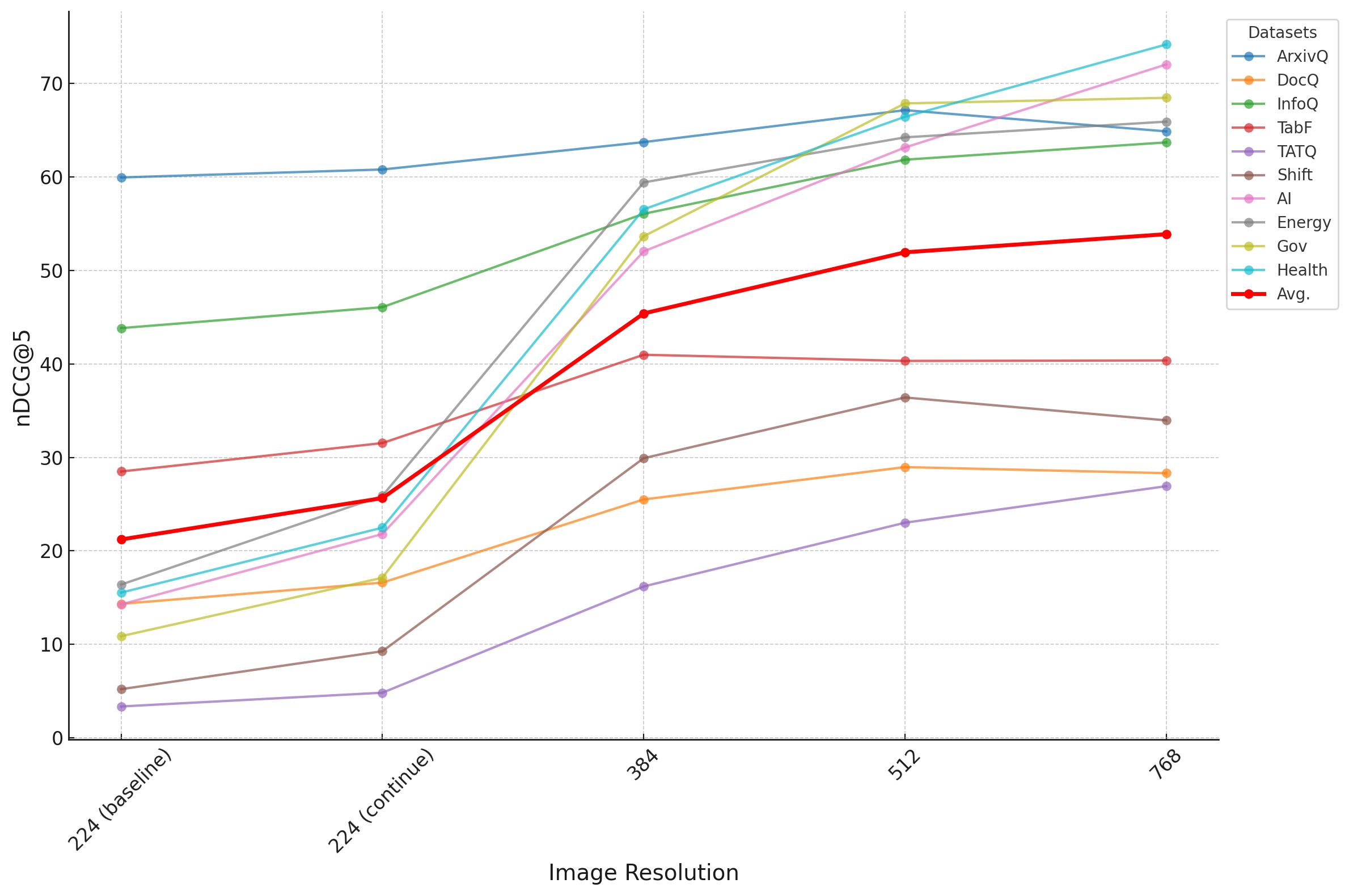}
    \caption{Performance on the ViDoRe benchmark \citep{colpali} against input resolution}
    \label{fig:resolution}
\end{figure}
Our analysis of performance at various stages of training demonstrates that image resolution plays a critical role in retrieving visually rich documents.
To better understand the relationship between image resolution and model performance on such documents, we conduct a targeted experiment.

We checkpointed \jclipII{} at the end of Stage 1, when it had only seen images with a resolution of $\mathrm{(224, 224)}$.
We conducted four runs, keeping the image resolution to $\mathrm{(224, 224)}$ for the first run and increasing to $\mathrm{(384, 384)}$, $\mathrm{(512, 512)}$, and $\mathrm{(768, 768)}$ for three additional runs.
Each run was trained for 3500 steps using the same visually rich training set and hyperparameters. The models were then evaluated on the ViDoRe benchmark \citep{colpali}, which comprises 10 datasets designed for retrieving visually rich documents with text queries. We plot the results in Figure \ref{fig:resolution}.

Unsurprisingly, increasing image resolution has a positive impact on linking queries to visually rich documents. The most significant improvement occurs when resolution increases from $\mathrm{(224, 224)}$ to $\mathrm{(384, 384)}$, with the average nDCG@5 score across 10 benchmarks rising from $\mathrm{0.256}$ to $\mathrm{0.454}$. A further increase in resolution to $\mathrm{(512, 512)}$ also brings a noticeable gain in performance.

Considering that the model is trained with a patch size of 14, increasing the resolution beyond $\mathrm{(512, 512)}$ raises the number of patches quadratically, for example, by a factor of $\mathrm{2.25x}$ when increasing the resolution to $\mathrm{(768, 768)}$. This large increase in computing costs yields only marginal improvements, with nDCG@5 increasing by just $\mathrm{0.019}$. We consider $\mathrm{(512, 512)}$ the optimal image resolution for this model, with a reasonable balance between performance and cost.

\subsection{Unified Batch Training vs Multi-Task Learning}

Our training approach, as outlined in Section \ref{sec:training}, involves two tasks at each stage, both optimized in a multi-task learning framework: a multimodal alignment task for images and text, and a second text-only task.
Both use the InfoNCE \citep{infonce} objective. Regardless of whether the loss is computed on image-text or text-text pairs, the same loss function is applied to pairs of vectors. This raises the question: Is there any value in merging the image-text pair and text-text pair data into mixed modality batches and optimizing a single contrastive objective?

We refer to this approach as the \textit{Unified Batch Training} technique, in contrast to the \textit{Multi-Task Learning} paradigm.
The motivation is two-fold: simplicity of computation, and the possibility of mitigating the modality gap \citep{modgap} by using in-batch negatives across modalities. Compared to multi-task training, the unified batch approach would use a single InfoNCE loss with a shared temperature value, attempting to force all modalities to align within the same embedding space. For this investigation, we also integrated a SimCLR-like self-supervised learning method \citep{simclr} into the contrastive training, aiming to provide more robust image representations. A schematic representation of the unified batch technique is given in Figure~\ref{appendix:unified-batch} in Section~\ref{appendix}.

Our Stage 1 experiments show that this technique does bring some early improvements, but fails beyond a certain point, at least with regard to crossmodal tasks. This was reflected in the slow decay of the loss temperature $\tau$ (our temperature is a trainable parameter in this case) and its eventual stabilization around $\mathrm{0.02}$. This is in contrast to the multi-task learning approach, where the temperature value decreases more rapidly and keeps decreasing throughout training down to $\mathrm{0.015}$, leading to better performance on crossmodal tasks.

We hypothesize that this limitation arises from the fundamental information asymmetry between the visual and textual modalities \citep{information_imbalance_clip}. Images are very information-dense, full of small details, while the corresponding textual descriptions are far less detailed \citep{information_imbalance_clip, rethinking_IR}. In temperature-scaled contrastive learning, the temperature parameter controls the hardness of the loss function with respect to the negative samples, i.e., the level of penalties on the hard negative samples \citep{understanding_contrastive}. From an information-theory perspective, the level of penalty on the hard negative text-text samples theoretically differs from that of image-text samples due to the information gap problem. 
Consequently, as supported by our experiments, enforcing a unified temperature parameter across both modalities is empirically suboptimal.

\section{Conclusion}
\label{sec:conclusion}

This work presents an enhanced strategy for training dual-encoder vision-language embedding models for multilingual crossmodal as well as text retrieval and semantic text similarity tasks using contrastive learning. We introduce improvements to the training, namely support for multiple languages, Matryoshka Representation Learning,
and visually-rich document understanding. We have used these improvements to train and release a new model, \jclipII{}, with strong crossmodal and text-only performance on standard benchmarks.
Finally, we have presented in this paper analyses of two important considerations for CLIP models going forward: the effects of the modality gap and the role of the image resolution in understanding complex visual inputs.

\bibliography{iclr2025_conference}
\bibliographystyle{iclr2025_conference}

%%%%%%%%%%%%%%%%%%%%%%%%%%%%%%%%%%%%%%%%%%%%%%%%%%%%%%%%%%%%%%%%%%%%%%%%
%%%%%%%%%%%%%%%%%%%%%%%%%%%%% APPENDIX %%%%%%%%%%%%%%%%%%%%%%%%%%%%%%%%%
%%%%%%%%%%%%%%%%%%%%%%%%%%%%%%%%%%%%%%%%%%%%%%%%%%%%%%%%%%%%%%%%%%%%%%%%

\newpage

\appendix
\section{Appendix}
\label{appendix}

\begin{table}[h!]
\centering
\caption{Training settings on each stage}
\label{appendix:training-settings}
\vspace{0.1in}
\begin{small}
\begin{tabular}{l|rrr}
\toprule
\textbf{Parameter} & \textbf{Stage 1} & \textbf{Stage 2} & \textbf{Stage 3} \\
\midrule
Image encoder weights init & EVA02 ViT L/14 \citep{evaclip} & Stage 1 & Stage 2 \\
Text encoder weights init & JinaXLMRoBERTa \citep{jembeddings3} & Stage 1 & Stage 2  \\
Peak image encoder LR & 2e-4 & 5e-5 & 5e-6 \\
Image encoder layer-wise LR decay & 1 & 0.98 & 1 \\
Peak text encoder LR & 1e-4 & 5e-5 & 1e-4 \\
Text encoder layer-wise LR decay & 1 & 0.98 & 1 \\
Image-text pairs batch size & $16,384$ & $8,192$ & $1,024$ \\
Text pairs batch size &  $16,384$ & $8,192$ & $128$ \\
Total steps & $100,000$ & $6,000$ & $16,000$ \\
Max sequence length & $77$ & $512$ & $512$ \\
Image-text pairs samples seen & 1.7B & 50M & 16M \\
Text pairs samples seen & 1.7B & 50M & 2M \\
Number of GPUs - H100s 80GB & 8 & 8 & 8 \\
Input resolution & $\mathrm{(224, 224)} \rightarrow \mathrm{(384, 384)}$ & $\mathrm{(384, 384)}$ & $\mathrm{(512, 512)}$ \\
Patch size & \multicolumn{3}{c}{$\mathrm{(14, 14)}$} \\
LR schedule & \multicolumn{3}{c}{cosine decay} \\
Optimizer & \multicolumn{3}{c}{AdamW \citep{adamw}} \\
Optimizer hyper-parameters & \multicolumn{3}{c}{\(\beta_1, \beta_2, \epsilon = 0.9, 0.98, 1e-6\)} \\
Weight decay & \multicolumn{3}{c}{0.02} \\
Numerical precision & \multicolumn{3}{c}{bfloat16} \\
\bottomrule
\end{tabular}
\end{small}
%}
\end{table}

\begin{table*}[htb]
    \centering
    \setlength{\tabcolsep}{4.5pt} 
    \caption{Model performance on the CLIP Benchmark retrieval tasks}
    \label{appendix:clipb}
    \vspace{0.1in}
    \begin{center}
    \begin{small}
    \begin{tabular}{l|cccccc}
        \toprule
        Dataset - Model & \makecell{jina-clip-v2} & \makecell{jina-clip-v2 \\ stage 1} & \makecell{jina-clip-v2 \\ stage 2} & \makecell{jina-clip-v1} & \makecell{nllb-siglip \\ large} &  \makecell{nllb-siglip \\ base} \\
        \midrule
        \multicolumn{7}{c}{\textbf{Zero-shot Image Retrieval - Recall@5  [\%]}} \\
        \midrule
        Flickr30K \citep{flickr30k} & 89.84 & 86.84 & 90.04 & 89.04 & \textbf{92.24} & 90.02 \\
        MS COCO \citep{mscococaptions}  & 68.35 & 60.91 & 69.59 & 66.42 & \textbf{70.84} & 69.13 \\
        \midrule
        \multicolumn{7}{c}{\textbf{Zero-shot Text Retrieval - Recall@5  [\%]}} \\
        \midrule
        Flickr30K \citep{flickr30k} & \textbf{98.00} & 96.10 & 97.40 & 96.40 & 97.10 & 95.00 \\
        MS COCO \citep{mscococaptions} &  81.46 & 77.12 & \textbf{81.74} & 79.02 & 79.20 & 77.76 \\
        \bottomrule
    \end{tabular}
\end{small}
\end{center}
\end{table*}

\begin{table*}[htb]
    \centering
    \setlength{\tabcolsep}{4.5pt} 
    \caption{Model performance on Crossmodal-3600 \citep{crossmodal3600}}
    \label{appendix:crossmodal3600}
    \vspace{0.1in}
    \begin{center}
    \begin{small}
    \begin{tabular}{l|cccccc}
    \toprule
    Language - Model & \makecell{jina-clip-v2} & \makecell{jina-clip-v2 \\ stage 1} & \makecell{jina-clip-v2 \\ stage 2}& \makecell{jina-clip-v1} & \makecell{nllb-siglip \\ large} &  \makecell{nllb-siglip \\ base} \\
    \midrule
    \multicolumn{7}{c}{\textbf{Zero-shot Image Retrieval - Recall@5  [\%]}} \\
    \midrule
    \textbf{average} & 81.43 & 73.51 & \textbf{84.13} & 16.93 & 82.07 & 79.29 \\
    ar & 73.56 & 66.22 & 76.89 & 0.19 & \textbf{78.92} & 76.94 \\
    bn & 63.78 & 50.19 & 68.11 & 0.11 & \textbf{75.19} & 74.19 \\
    da & 85.39 & 76.53 & \textbf{87.97} & 15.39 & 87.14 & 86.64 \\
    de & 91.25 & 84.64 & \textbf{92.42} & 37.42 & 89.56 & 87.25 \\
    el & 75.03 & 68.53 & \textbf{77.92} & 0.56 & 77.83 & 71.97 \\
    en & 75.83 & 69.78 & \textbf{77.78} & 76.17 & 73.11 & 72.22 \\
    es & 83.64 & 78.28 & \textbf{86.28} & 47.36 & 82.64 & 79.97 \\
    fi & 82.83 & 75.28 & 85.89 & 5.03 & \textbf{86.42} & 81.44 \\
    fr & 88.78 & 83.50 & \textbf{90.67} & 62.33 & 87.86 & 85.58 \\
    hi & 55.25 & 42.03 & 59.47 & 0.11 & \textbf{60.31} & 58.08 \\
    id & 84.22 & 73.69 & \textbf{86.94} & 13.14 & 86.31 & 84.17 \\
    it & 88.33 & 81.19 & \textbf{89.64} & 33.39 & 85.94 & 82.67 \\
    ja & 87.03 & 77.28 & \textbf{90.03} & 3.75 & 86.06 & 83.14 \\
    ko & 78.81 & 71.81 & \textbf{83.22} & 0.33 & 78.75 & 75.47 \\
    nl & 82.56 & 75.72 & \textbf{84.47} & 27.69 & 81.69 & 78.86 \\
    no & 81.08 & 71.97 & \textbf{83.08} & 15.61 & 82.69 & 79.97 \\
    pl & 84.00 & 79.33 & \textbf{86.50} & 7.39 & 82.72 & 78.61 \\
    pt & 82.42 & 76.17 & \textbf{85.19} & 31.97 & 82.69 & 79.44 \\
    ro & 89.36 & 82.92 & \textbf{92.22} & 17.89 & 90.03 & 86.17 \\
    ru & 88.97 & 82.97 & \textbf{91.11} & 2.19 & 86.44 & 83.78 \\
    sv & 78.06 & 71.33 & \textbf{80.56} & 15.22 & 79.33 & 76.17 \\
    th & 81.61 & 70.92 & \textbf{85.08} & 1.89 & 81.14 & 78.83 \\
    tr & 81.31 & 75.31 & \textbf{84.53} & 4.86 & 83.47 & 81.00 \\
    uk & 88.56 & 82.28 & \textbf{89.89} & 0.94 & 85.44 & 81.89 \\
    vi & 86.64 & 76.31 & \textbf{89.06} & 3.17 & 85.56 & 82.56 \\
    zh & 78.97 & 66.97 & \textbf{82.50} & 2.06 & 76.56 & 74.64 \\
    \midrule
    \multicolumn{7}{c}{\textbf{Zero-shot Text Retrieval - Recall@5  [\%]}} \\
    \midrule
    \textbf{average} & 83.23 & 79.64 & \textbf{84.12} & 19.82 & 80.16 & 76.56 \\
    ar & 76.25 & 72.14 & \textbf{76.67} & 0.31 & 75.86 & 73.69 \\
    bn & 69.00 & 63.78 & 70.22 & 0.11 & \textbf{75.58} & 73.61 \\
    da & 88.53 & 84.42 & \textbf{88.86} & 17.25 & 86.53 & 84.69 \\
    de & \textbf{92.47} & 88.42 & \textbf{92.47} & 39.75 & 87.50 & 84.44 \\
    el & 73.33 & 73.22 & \textbf{75.61} & 0.64 & 74.81 & 68.89 \\
    en & 78.58 & 74.33 & \textbf{79.61} & 79.06 & 70.81 & 69.08 \\
    es & 86.28 & 81.19 & \textbf{87.36} & 49.14 & 81.19 & 77.00 \\
    fi & 84.19 & 81.17 & \textbf{85.64} & 7.22 & 84.25 & 78.75 \\
    fr & 90.89 & 86.56 & \textbf{91.00} & 63.14 & 86.64 & 83.64 \\
    hi & 61.64 & 54.39 & 61.56 & 0.08 & \textbf{61.89} & 59.83 \\
    id & 86.31 & 83.64 & \textbf{87.64} & 16.75 & 84.33 & 81.89 \\
    it & 90.17 & 84.31 & \textbf{90.53} & 36.53 & 83.50 & 78.64 \\
    ja & 88.50 & 85.53 & \textbf{89.44} & 7.69 & 84.03 & 81.14 \\
    ko & 81.42 & 79.22 & \textbf{83.00} & 0.53 & 76.75 & 74.17 \\
    nl & 82.47 & 77.94 & \textbf{83.33} & 29.28 & 79.28 & 74.33 \\
    no & 83.75 & 77.22 & \textbf{84.42} & 18.53 & 82.08 & 77.03 \\
    pl & 84.61 & 83.72 & \textbf{85.69} & 10.00 & 79.58 & 75.39 \\
    pt & 83.94 & 79.31 & \textbf{84.06} & 33.33 & 79.39 & 74.89 \\
    ro & 91.31 & 87.39 & \textbf{92.08} & 18.83 & 88.83 & 83.58 \\
    ru & 90.64 & 87.75 & \textbf{90.89} & 3.33 & 84.64 & 80.61 \\
    sv & 78.28 & 77.53 & \textbf{80.11} & 16.50 & 76.58 & 72.31 \\
    th & 81.94 & 80.22 & \textbf{84.11} & 2.17 & 79.56 & 76.28 \\
    tr & 82.67 & 80.36 & \textbf{83.81} & 6.94 & 80.36 & 77.69 \\
    uk & 89.53 & 86.42 & \textbf{89.97} & 1.58 & 83.22 & 78.67 \\
    vi & 88.06 & 85.64 & \textbf{88.58} & 6.06 & 83.14 & 80.03 \\
    zh & 79.22 & 74.94 & \textbf{80.42} & 4.67 & 73.83 & 70.22 \\
    \bottomrule
    \end{tabular}
\end{small}
\end{center}
\end{table*}

\begin{table*}[htb]
    \centering
    \setlength{\tabcolsep}{4.5pt} 
    \caption{Model performance on XTD10 \citep{xtd10, mic}}
    \label{appendix:xtd10}
    \vspace{0.1in}
    \begin{center}
    \begin{small}
    \begin{tabular}{l|cccccc}
    \toprule
    Language - Model & \makecell{jina-clip-v2} & \makecell{jina-clip-v2 \\ stage 1} & \makecell{jina-clip-v2 \\ stage 2} & \makecell{jina-clip-v1} & \makecell{nllb-siglip \\ large} &  \makecell{nllb-siglip \\ base} \\
    \midrule
    \multicolumn{7}{c}{\textbf{Zero-shot Image Retrieval - Recall@5  [\%]}} \\
    \midrule
    \textbf{average} & 84.87 & 80.66 & 86.11 & 31.01 & \textbf{87.60} & 86.23 \\
    de & 85.70 & 80.40 & 86.80 & 48.80 & \textbf{88.30} & 87.00 \\
    en & \textbf{89.40} & 84.00 & \textbf{89.40} & 89.00 & \textbf{89.40} & 88.80 \\
    es & 85.90 & 82.70 & 87.40 & 56.80 & \textbf{88.20} & 86.70 \\
    fr & 85.10 & 80.90 & 87.30 & 66.80 & \textbf{87.70} & 86.90 \\
    it & 85.80 & 83.20 & 87.10 & 45.00 & \textbf{89.30} & 87.80 \\
    ko & 82.10 & 78.30 & 83.00 & 1.30 & \textbf{85.20} & 83.60 \\
    pl & 86.50 & 81.10 & 88.00 & 11.00 & \textbf{89.40} & 87.60 \\
    ru & 81.10 & 79.30 & 82.10 & 3.20 & \textbf{83.40} & 82.40 \\
    tr & 83.70 & 81.30 & 86.30 & 7.00 & \textbf{88.30} & 86.80 \\
    zh & 83.40 & 75.40 & 83.70 & 4.50 & \textbf{86.80} & 84.70 \\
    \midrule
    \multicolumn{7}{c}{\textbf{Zero-shot Text Retrieval - Recall@5  [\%]}} \\
    \midrule
    \textbf{average} & 86.03 & 83.02 & \textbf{86.45} & 36.89 & 85.37 & 84.56 \\
    de & 86.20 & 82.90 & \textbf{86.90} & 50.10 & 84.40 & 84.30 \\
    en & 90.50 & 87.80 & \textbf{90.60} & 89.40 & 88.30 & 87.50 \\
    es & 87.00 & 84.90 & \textbf{87.50} & 55.60 & 85.80 & 85.20 \\
    fr & 85.20 & 82.20 & 85.00 & 67.30 & \textbf{86.30} & 84.40 \\
    it & \textbf{88.00} & 83.10 & 87.70 & 51.10 & 86.70 & 85.60 \\
    ko & 82.10 & 79.30 & \textbf{84.00} & 1.60 & 83.20 & 82.60 \\
    pl & 88.50 & 84.50 & \textbf{88.90} & 15.50 & 86.80 & 86.60 \\
    ru & \textbf{83.10} & 79.60 & 81.80 & 3.50 & 80.90 & 81.20 \\
    tr & 85.60 & 83.20 & 85.90 & 9.50 & \textbf{87.10} & 85.40 \\
    zh & 84.10 & 82.70 & \textbf{86.20} & 7.40 & 84.20 & 82.80 \\
    \bottomrule
    \end{tabular}
\end{small}
\end{center}
\end{table*}

\begin{table*}[ht]
  \centering
  \caption{Retrieval and STS evaluation on English MTEB \citep{mteb} (1/3)}
  \label{appendix:results-mteb-en-1}
  \setlength{\tabcolsep}{4.5pt}
  \vskip 0.1in
  \tiny{
    \begin{tabular}{cc|ccccccc}
    \toprule
    Task & Split & \texttt{{jc-v1}} & \texttt{{nllb-b}} & \texttt{{nllb-l}} & \texttt{{je-v3}} & \texttt{{jc-v2-s1}} & \texttt{{jc-v2-s2}} & \texttt{{jc-v2}} \\
    \midrule
    \midrule
    \multicolumn{9}{c}{\textbf{Retrieval} nDCG@10 [\%]} \\
    \midrule
    AILACasedocs & test & 32.79 & 6.62 & 8.63 & \textbf{34.73} & 29.21 & 31.88 & 23.10 \\
    AILAStatutes & test & 14.45 & 15.04 & 13.58 & \textbf{33.00} & 22.58 & 26.45 & 8.86 \\
    ARCChallenge & test & 10.52 & 5.85 & 6.78 & 10.12 & \textbf{11.18} & 10.92 & 10.96 \\
    AlphaNLI & test & \textbf{31.45} & 16.02 & 19.25 & 30.16 & 28.39 & 27.87 & 29.12 \\
    ArguAna & test & \textbf{49.41} & 25.01 & 33.81 & 43.28 & 42.85 & 47.99 & 43.65 \\
    BelebeleRetrieval & test & 91.83 & 72.88 & 77.43 & \textbf{93.25} & 91.84 & 91.86 & 92.97 \\
    \makecell{CQADupstack \\ AndroidRetrieval} & test & 52.14 & 27.46 & 31.30 & 37.01 & 52.14 & \textbf{52.58} & 50.21 \\
    \makecell{CQADupstack \\ EnglishRetrieval} & test & 45.74 & 19.98 & 23.40 & 44.50 & \textbf{48.67} & 48.19 & 46.36 \\
    \makecell{CQADupstack \\ GamingRetrieval} & test & \textbf{59.61} & 34.23 & 37.64 & 47.53 & 58.56 & 58.39 & 54.40 \\
    \makecell{CQADupstack \\ GisRetrieval} & test & 39.30 & 17.70 & 20.62 & 35.66 & 38.72 & 39.48 & \textbf{40.61} \\
    \makecell{CQADupstack \\ MathematicaRetrieval} & test & 28.23 & 13.08 & 15.66 & 30.23 & 30.69 & \textbf{31.37} & 30.70 \\
    \makecell{CQADupstack \\ PhysicsRetrieval} & test & 46.31 & 25.37 & 28.74 & 42.18 & 45.62 & \textbf{46.61} & 45.84 \\
    \makecell{CQADupstack \\ ProgrammersRetrieval} & test & 41.40 & 21.06 & 25.66 & 37.33 & 41.32 & \textbf{42.61} & 41.30 \\
    CQADupstackRetrieval & test & 40.93 & 20.74 & 23.54 & 36.02 & 41.48 & \textbf{42.17} & 41.19 \\
    \makecell{CQADupstack \\ StatsRetrieval} & test & 33.97 & 19.10 & 20.63 & 31.20 & 35.06 & \textbf{35.43} & 35.41 \\
    \makecell{CQADupstack \\ TexRetrieval} & test & 29.43 & 12.42 & 13.46 & 26.97 & 29.71 & 30.41 & \textbf{30.92} \\
    \makecell{CQADupstack \\ UnixRetrieval} & test & 41.13 & 19.57 & 20.99 & 37.22 & 42.60 & \textbf{43.01} & 42.48 \\
    \makecell{CQADupstack \\ WebmastersRetrieval} & test & 41.43 & 23.31 & 26.58 & 33.16 & 41.81 & \textbf{43.94} & 42.01 \\
    \makecell{CQADupstack \\ WordpressRetrieval} & test & 32.43 & 15.65 & 17.87 & 29.20 & 32.92 & 34.05 & \textbf{34.05} \\
    CUREv1 & dentistry\_and\_oral\_health & 45.53 & 20.08 & 26.11 & 49.19 & 41.41 & 40.88 & \textbf{50.26} \\
    CUREv1 & dermatology & 51.32 & 17.25 & 22.31 & 53.54 & 41.41 & 42.34 & \textbf{59.59} \\
    CUREv1 & gastroenterology & 45.20 & 19.99 & 23.25 & 49.36 & 40.56 & 41.59 & \textbf{51.96} \\
    CUREv1 & genetics & 48.26 & 18.91 & 27.19 & \textbf{56.64} & 41.62 & 43.95 & 54.23 \\
    CUREv1 & neuroscience\_and\_neurology & 38.67 & 15.97 & 21.31 & \textbf{44.91} & 35.35 & 35.65 & 42.92 \\
    CUREv1 & orthopedic\_surgery & 45.13 & 12.79 & 19.48 & 44.11 & 40.72 & 38.91 & \textbf{46.29} \\
    CUREv1 & otorhinolaryngology & 39.65 & 12.28 & 18.19 & 41.91 & 34.29 & 32.90 & \textbf{46.59} \\
    CUREv1 & plastic\_surgery & 46.10 & 15.43 & 23.43 & 46.86 & 40.10 & 40.62 & \textbf{48.34} \\
    CUREv1 & psychiatry\_and\_psychology & 46.87 & 21.02 & 27.88 & \textbf{52.34} & 41.78 & 42.10 & 50.63 \\
    CUREv1 & pulmonology & 45.08 & 16.75 & 26.42 & \textbf{53.72} & 41.09 & 40.11 & 51.80 \\
    CUREv1 & avg & 45.18 & 17.05 & 23.56 & 49.26 & 39.83 & 39.91 & \textbf{50.26} \\
    \makecell{ChemHotpotQA \\ Retrieval} & test & \textbf{78.05} & 41.81 & 44.20 & 68.65 & 63.21 & 61.77 & 58.96 \\
    ChemNQRetrieval & test & 57.11 & 39.22 & 39.19 & \textbf{60.80} & 48.21 & 47.78 & 54.58 \\
    ClimateFEVER & test & 24.82 & 13.06 & 16.44 & \textbf{42.26} & 19.31 & 23.70 & 30.14 \\
    \makecell{ClimateFEVER \\ HardNegatives} & test & 25.32 & 14.03 & 17.11 & \textbf{43.02} & 20.16 & 24.22 & 30.47 \\
    CodeFeedbackMT & test & 39.15 & 4.86 & 6.66 & \textbf{59.84} & 50.42 & 52.46 & 47.11 \\
    CodeFeedbackST & test & 66.06 & 17.09 & 20.64 & \textbf{78.09} & 70.81 & 71.05 & 68.62 \\
    DBPedia & test & 36.66 & 17.76 & 21.37 & \textbf{41.00} & 22.03 & 24.55 & 37.42 \\
    DBPediaHardNegatives & test & 38.84 & 23.32 & 26.66 & \textbf{43.64} & 26.47 & 28.54 & 40.76 \\
    FEVER & test & 76.30 & 29.84 & 33.91 & \textbf{89.07} & 60.54 & 67.09 & 84.98 \\
    FEVERHardNegatives & test & 77.02 & 38.27 & 40.69 & \textbf{89.86} & 61.52 & 68.72 & 86.56 \\
    FaithDial & test & 23.18 & 16.83 & 18.69 & \textbf{26.35} & 24.03 & 23.93 & 22.84 \\
    FeedbackQARetrieval & test & 49.15 & 22.69 & 33.43 & \textbf{52.06} & 45.38 & 49.90 & 51.00 \\
    FiQA2018 & test & 38.27 & 8.74 & 12.79 & \textbf{47.46} & 41.65 & 40.96 & 42.00 \\
    HagridRetrieval & dev & \textbf{98.69} & 96.85 & 97.16 & 98.69 & 98.65 & 98.42 & 98.69 \\
    HellaSwag & test & 27.30 & 14.33 & 18.39 & 29.26 & \textbf{29.32} & 29.02 & 27.21 \\
    HotpotQA & test & 61.89 & 21.55 & 23.09 & \textbf{64.65} & 49.63 & 46.96 & 60.15 \\
    \makecell{HotpotQA \\ HardNegatives} & test & 62.58 & 27.56 & 28.59 & \textbf{64.74} & 52.00 & 49.67 & 60.01 \\
    \makecell{LEMBNarrative \\ QARetrieval} & test & 33.55 & 10.82 & 12.62 & \textbf{34.22} & 19.86 & 19.68 & 23.57 \\
    LEMBNeedleRetrieval & test\_256 & 72.00 & 36.00 & 40.00 & 64.00 & 78.00 & 72.00 & \textbf{86.00} \\
    LEMBNeedleRetrieval & test\_512 & 58.00 & 24.00 & 18.00 & 32.00 & 64.00 & 54.00 & \textbf{70.00} \\
    LEMBNeedleRetrieval & test\_1024 & \textbf{66.00} & 6.00 & 6.00 & 14.00 & 64.00 & 50.00 & 46.00 \\
    LEMBNeedleRetrieval & test\_2048 & \textbf{76.00} & 8.00 & 8.00 & 10.00 & 60.00 & 56.00 & 58.00 \\
    LEMBNeedleRetrieval & test\_4096 & \textbf{66.00} & 2.00 & 2.00 & 8.00 & 48.00 & 48.00 & 54.00 \\
    LEMBNeedleRetrieval & test\_8192 & \textbf{72.00} & 0.00 & 2.00 & 12.00 & 46.00 & 46.00 & 34.00 \\
    LEMBNeedleRetrieval & test\_16384 & \textbf{42.00} & 0.00 & 0.00 & 8.00 & 28.00 & 26.00 & 28.00 \\
    LEMBNeedleRetrieval & test\_32768 & \textbf{16.00} & 2.00 & 4.00 & 2.00 & 6.00 & 8.00 & 8.00 \\
    LEMBNeedleRetrieval & avg & \textbf{58.50} & 9.75 & 10.00 & 18.75 & 49.25 & 45.00 & 48.00 \\
    LEMBPasskeyRetrieval & test\_256 & 90.00 & 22.00 & 24.00 & \textbf{100.00} & 98.00 & 76.00 & 100.00 \\
    LEMBPasskeyRetrieval & test\_512 & 52.00 & 20.00 & 16.00 & \textbf{100.00} & 58.00 & 72.00 & 98.00 \\
    LEMBPasskeyRetrieval & test\_1024 & 34.00 & 14.00 & 14.00 & \textbf{94.00} & 16.00 & 42.00 & 56.00 \\
    \bottomrule
    \end{tabular}
    \\
    \vspace{0.1in}
    \footnotesize{Evaluated on all tasks included in MTEB version $\mathrm{1.34.7}$, except the following: \texttt{MrTidyRetrieval}, \texttt{BrightRetrieval}, \texttt{MSMARCOv2}, \texttt{NeuCLIR2022Retrieval}, \texttt{NeuCLIR2023Retrieval} and \texttt{MIRACLRetrieval}. These tasks were excluded either due to bugs in the evaluation code or excessive computation times.} \\
  }
\end{table*}

\begin{table*}[ht]
  \centering
  \caption{Retrieval and STS evaluation on English MTEB \citep{mteb} (2/3)}
  \label{appendix:results-mteb-en-2}
  \setlength{\tabcolsep}{4.5pt}
  \vskip 0.1in
  \tiny{
    \begin{tabular}{cc|ccccccc}
    \toprule
    Task & Split & \texttt{{jc-v1}} & \texttt{{nllb-b}} & \texttt{{nllb-l}} & \texttt{{je-v3}} & \texttt{{jc-v2-s1}} & \texttt{{jc-v2-s2}} & \texttt{{jc-v2}} \\
    \midrule
    LEMBPasskeyRetrieval & test\_2048 & 34.00 & 4.00 & 4.00 & \textbf{72.00} & 12.00 & 14.00 & 12.00 \\
    LEMBPasskeyRetrieval & test\_4096 & \textbf{72.00} & 0.00 & 0.00 & 42.00 & 2.00 & 20.00 & 0.00 \\
    LEMBPasskeyRetrieval & test\_8192 & \textbf{82.00} & 0.00 & 2.00 & 26.00 & 10.00 & 4.00 & 8.00 \\
    LEMBPasskeyRetrieval & test\_16384 & \textbf{44.00} & 2.00 & 2.00 & 26.00 & 8.00 & 4.00 & 6.00 \\
    LEMBPasskeyRetrieval & test\_32768 & \textbf{26.00} & 2.00 & 0.00 & 10.00 & 6.00 & 8.00 & 4.00 \\
    LEMBPasskeyRetrieval & avg & 54.25 & 8.00 & 7.75 & \textbf{58.75} & 26.25 & 30.00 & 35.50 \\
    LEMBQMSumRetrieval & test & 37.38 & 8.05 & 9.57 & \textbf{39.35} & 29.37 & 29.93 & 31.07 \\
    \makecell{LEMBSummScreen \\ FDRetrieval} & validation & \textbf{93.37} & 23.44 & 23.31 & 92.22 & 55.86 & 56.84 & 75.02 \\
    LEMBWikimQARetrieval & test & \textbf{75.19} & 33.00 & 35.49 & 65.95 & 57.39 & 55.88 & 64.85 \\
    \makecell{LegalBench \\ ConsumerContractsQA} & test & 69.55 & 39.91 & 44.37 & \textbf{78.14} & 61.34 & 67.18 & 64.18 \\
    \makecell{LegalBench \\ CorporateLobbying} & test & 89.09 & 81.55 & 86.17 & \textbf{93.65} & 88.40 & 90.37 & 91.57 \\
    LegalSummarization & test & \textbf{63.23} & 45.25 & 50.57 & 59.25 & 53.08 & 56.06 & 62.70 \\
    LitSearchRetrieval & test & 41.89 & 11.96 & 12.94 & 48.18 & 39.51 & 39.15 & \textbf{48.37} \\
    \makecell{MIRACLRetrieval \\ HardNegatives} & dev & 43.34 & 19.57 & 20.44 & \textbf{51.97} & 31.97 & 39.01 & 50.77 \\
    MLQARetrieval & test & 63.42 & 35.82 & 40.61 & \textbf{64.72} & 59.87 & 60.03 & 64.08 \\
    MLQuestions & test & 59.43 & 27.79 & 30.23 & \textbf{63.10} & 54.91 & 54.74 & 56.11 \\
    MSMARCO & dev & 36.92 & 10.63 & 13.80 & \textbf{40.84} & 22.94 & 27.12 & 37.40 \\
    MSMARCOHardNegatives & test & 65.43 & 41.81 & 44.71 & \textbf{71.74} & 55.21 & 58.27 & 69.89 \\
    MedicalQARetrieval & test & 67.24 & 34.04 & 39.22 & 70.23 & 66.01 & 68.24 & \textbf{71.34} \\
    \makecell{MultiLongDoc \\ Retrieval} & test & \textbf{34.78} & 8.38 & 10.80 & 28.87 & 19.50 & 18.97 & 20.16 \\
    NFCorpus & test & 33.52 & 16.77 & 22.13 & \textbf{36.61} & 31.00 & 32.16 & 32.88 \\
    NQ & test & 58.11 & 16.28 & 19.94 & \textbf{64.33} & 40.00 & 43.03 & 57.16 \\
    NQHardNegatives & test & 59.43 & 21.05 & 23.33 & \textbf{65.06} & 41.42 & 44.50 & 58.50 \\
    NanoArguAnaRetrieval & train & 55.13 & 30.90 & 37.50 & 49.81 & 49.93 & \textbf{59.43} & 51.27 \\
    \makecell{NanoClimateFever \\ Retrieval} & train & 29.09 & 21.87 & 27.72 & \textbf{44.57} & 28.17 & 29.40 & 34.37 \\
    NanoDBPediaRetrieval & train & 57.42 & 40.98 & 41.53 & \textbf{61.81} & 49.16 & 53.60 & 59.45 \\
    NanoFEVERRetrieval & train & 88.59 & 56.74 & 57.84 & \textbf{91.93} & 80.95 & 82.60 & 90.54 \\
    \makecell{NanoFiQA2018 \\ Retrieval} & train & 46.25 & 17.92 & 26.05 & 53.61 & \textbf{53.80} & 53.22 & 53.41 \\
    \makecell{NanoHotpot \\ QARetrieval} & train & 75.15 & 42.22 & 51.20 & \textbf{75.65} & 67.68 & 65.87 & 72.86 \\
    NanoMSMARCORetrieval & train & 61.17 & 30.93 & 41.23 & \textbf{66.53} & 48.93 & 53.50 & 61.97 \\
    \makecell{NanoNFCorpus \\ Retrieval} & train & 34.05 & 16.33 & 20.63 & \textbf{36.85} & 29.81 & 31.01 & 32.01 \\
    NanoNQRetrieval & train & 69.86 & 28.61 & 36.17 & \textbf{71.94} & 61.82 & 57.89 & 66.80 \\
    NanoQuoraRetrieval & train & 94.66 & 92.18 & 89.85 & 76.55 & 94.22 & 93.80 & \textbf{95.87} \\
    NanoSCIDOCSRetrieval & train & 38.47 & 21.36 & 28.36 & 40.42 & \textbf{41.56} & 41.27 & 39.07 \\
    NanoSciFactRetrieval & train & 71.75 & 24.97 & 33.39 & \textbf{79.08} & 71.41 & 72.98 & 70.20 \\
    \makecell{NanoTouche2020 \\ Retrieval} & train & 50.02 & 30.00 & 36.17 & \textbf{53.24} & 45.07 & 48.62 & 52.28 \\
    NarrativeQARetrieval & test & 33.45 & 11.31 & 12.58 & \textbf{34.13} & 19.83 & 19.63 & 23.51 \\
    PIQA & test & 30.40 & 12.58 & 15.64 & \textbf{32.02} & 28.67 & 30.35 & 30.65 \\
    PublicHealthQA & test & 79.51 & 56.80 & 64.25 & 83.56 & 81.30 & \textbf{84.56} & 84.12 \\
    Quail & test & \textbf{4.38} & 1.51 & 1.85 & 4.20 & 4.11 & 4.01 & 3.33 \\
    QuoraRetrieval & test & 87.88 & 77.12 & 80.47 & 61.68 & 87.14 & 86.98 & \textbf{88.14} \\
    \makecell{QuoraRetrieval \\ HardNegatives} & test & 87.88 & 77.08 & 80.41 & 63.39 & 87.39 & 87.24 & \textbf{87.98} \\
    RARbCode & test & 38.02 & 0.21 & 1.77 & \textbf{54.61} & 42.50 & 41.59 & 35.27 \\
    RARbMath & test & 53.29 & 14.87 & 18.83 & \textbf{74.57} & 61.83 & 62.81 & 59.99 \\
    SCIDOCS & test & 20.23 & 9.87 & 10.87 & 19.91 & 20.51 & \textbf{20.85} & 18.93 \\
    SCIDOCS-NL & test & 8.19 & 5.30 & 7.94 & \textbf{17.16} & 14.43 & 16.71 & 15.50 \\
    SIQA & test & 1.87 & 0.81 & 0.76 & 0.79 & 1.99 & 2.27 & \textbf{2.31} \\
    SciFact & test & 67.32 & 28.65 & 33.64 & \textbf{72.56} & 68.18 & 66.95 & 65.21 \\
    SpartQA & test & 7.67 & 0.67 & 3.22 & 0.73 & 0.10 & 1.82 & \textbf{8.51} \\
    StackOverflowQA & test & 82.28 & 14.30 & 19.87 & \textbf{90.79} & 81.33 & 86.46 & 84.43 \\
    \makecell{StatcanDialogue \\ DatasetRetrieval} & test & 26.33 & 14.73 & 15.49 & \textbf{33.44} & 4.26 & 11.41 & 26.09 \\
    TRECCOVID & test & 71.58 & 37.48 & 45.05 & \textbf{77.34} & 51.51 & 58.65 & 76.69 \\
    TempReasonL1 & test & \textbf{1.49} & 0.27 & 0.44 & 0.60 & 1.12 & 1.19 & 1.19 \\
    TempReasonL2Context & test & 9.15 & 5.95 & 7.88 & \textbf{11.73} & 6.85 & 6.78 & 7.82 \\
    TempReasonL2Fact & test & 15.07 & 6.26 & 7.92 & \textbf{19.70} & 11.98 & 11.72 & 16.03 \\
    TempReasonL2Pure & test & 0.95 & 0.12 & 0.23 & 0.50 & 0.96 & \textbf{1.14} & 1.03 \\
    TempReasonL3Context & test & 8.73 & 6.78 & 7.79 & \textbf{11.62} & 5.36 & 5.70 & 6.81 \\
    TempReasonL3Fact & test & 13.43 & 6.94 & 7.29 & \textbf{18.35} & 9.69 & 10.19 & 12.33 \\
    TempReasonL3Pure & test & \textbf{6.30} & 3.62 & 3.86 & 5.45 & 3.65 & 4.70 & 4.50 \\
    TopiOCQA & validation & 14.13 & 5.59 & 6.04 & \textbf{19.18} & 12.52 & 12.04 & 15.16 \\
    \makecell{TopiOCQA \\ HardNegatives} & validation & 14.02 & 6.17 & 6.98 & \textbf{19.19} & 13.23 & 12.58 & 15.03 \\
    \bottomrule
    \end{tabular}
    \\
    \vspace{0.1in}
    \footnotesize{Evaluated on all tasks included in MTEB version $\mathrm{1.34.7}$, except the following: \texttt{MrTidyRetrieval}, \texttt{BrightRetrieval}, \texttt{MSMARCOv2}, \texttt{NeuCLIR2022Retrieval}, \texttt{NeuCLIR2023Retrieval} and \texttt{MIRACLRetrieval}. These tasks were excluded either due to bugs in the evaluation code or excessive computation times.} \\
  }
\end{table*}

\begin{table*}[ht]
  \centering
  \caption{Retrieval and STS evaluation on English MTEB \citep{mteb} (3/3)}
  \label{appendix:results-mteb-en-3}
  \setlength{\tabcolsep}{4.5pt}
  \vskip 0.1in
  \tiny{
    \begin{tabular}{cc|ccccccc}
    \toprule
    Task & Split & \texttt{{jc-v1}} & \texttt{{nllb-b}} & \texttt{{nllb-l}} & \texttt{{je-v3}} & \texttt{{jc-v2-s1}} & \texttt{{jc-v2-s2}} & \texttt{{jc-v2}} \\
    \midrule
    \makecell{Touche2020 \\ Retrieval.v3} & test & \textbf{57.00} & 27.64 & 33.36 & 55.40 & 47.28 & 52.60 & 56.19 \\
    \makecell{WikipediaRetrieval \\ Multilingual} & test & 90.91 & 73.24 & 76.80 & 91.78 & 89.89 & 90.40 & \textbf{91.85} \\
    WinoGrande & test & \textbf{49.60} & 45.80 & 44.84 & 19.58 & 22.21 & 28.12 & 43.78 \\
    XMarket & test & 30.73 & 7.26 & 13.61 & \textbf{34.79} & 14.79 & 21.83 & 29.33 \\
    XQuADRetrieval & validation & 94.73 & 75.69 & 79.50 & \textbf{96.11} & 92.72 & 92.97 & 94.79 \\
    \midrule
    \multicolumn{2}{l|}{\textbf{Average}} & 47.04 & 24.69 & 27.83 & \textbf{48.34} & 42.02 & 43.47 & 46.46 \\
    \midrule
    \midrule
    \multicolumn{9}{c}{\textbf{STS} Spearman correlation on cosine similarity} \\
    \midrule
    BIOSSES & test & 83.75 & 65.77 & 60.63 & \textbf{84.52} & 83.53 & 83.14 & 82.90 \\
    SICK-R & test & 78.95 & 73.73 & 75.82 & 77.25 & 78.97 & 78.72 & \textbf{82.40} \\
    STS12 & test & 73.52 & 70.67 & 73.33 & \textbf{78.34} & 73.43 & 74.52 & 76.71 \\
    STS13 & test & 83.24 & 73.13 & 76.43 & \textbf{87.10} & 80.22 & 82.29 & 79.92 \\
    STS14 & test & 78.67 & 69.76 & 72.23 & \textbf{80.15} & 75.11 & 76.88 & 77.50 \\
    STS15 & test & \textbf{87.46} & 80.87 & 80.90 & 87.24 & 84.99 & 85.57 & 86.43 \\
    STS16 & test & 83.77 & 73.88 & 77.03 & 83.83 & 83.71 & 82.71 & \textbf{85.19} \\
    STS17 & test & \textbf{89.78} & 82.77 & 83.66 & 88.33 & 87.89 & 86.93 & 87.92 \\
    STS22.v2 & test & 65.84 & 55.21 & 58.06 & 65.50 & 63.46 & \textbf{68.19} & 67.00 \\
    STSBenchmark & test & 84.93 & 74.51 & 79.45 & 85.09 & 84.30 & 84.46 & \textbf{86.86} \\
    \makecell{STSBenchmark \\ MultilingualSTS} & test & 84.93 & 74.51 & 79.45 & 85.09 & 84.30 & 84.45 & \textbf{86.86} \\
    SemRel24STS & test & \textbf{81.36} & 76.07 & 79.75 & 77.70 & 80.19 & 81.29 & 79.31 \\
    \midrule
    \multicolumn{2}{l|}{\textbf{Average}} & 81.35 & 72.57 & 74.73 & \textbf{81.68} & 80.01 & 80.76 & 81.58 \\
    \midrule
    \bottomrule
    \end{tabular}
    \\
    \vspace{0.1in}
    \footnotesize{Evaluated on all tasks included in MTEB version $\mathrm{1.34.7}$, except the following: \texttt{MrTidyRetrieval}, \texttt{BrightRetrieval}, \texttt{MSMARCOv2}, \texttt{NeuCLIR2022Retrieval}, \texttt{NeuCLIR2023Retrieval} and \texttt{MIRACLRetrieval}. These tasks were excluded either due to bugs in the evaluation code or excessive computation times.} \\
  }
\end{table*}

\begin{table*}[ht]
  \centering
  \caption{Retrieval and STS evaluation on Chinese MTEB \citep{mteb} (1/1)}
  \label{appendix:results-mteb-zh-1}
  \setlength{\tabcolsep}{4.5pt}
  \vskip 0.1in
  \tiny{
    \begin{tabular}{cc|ccccccc}
    \toprule
    Task & Split & \texttt{{jc-v1}} & \texttt{{nllb-b}} & \texttt{{nllb-l}} & \texttt{{je-v3}} & \texttt{{jc-v2-s1}} & \texttt{{jc-v2-s2}} & \texttt{{jc-v2}} \\
    \midrule
    \midrule
    \multicolumn{9}{c}{\textbf{Retrieval} nDCG@10 [\%]} \\
    \midrule
    BelebeleRetrieval & test & 15.96 & 63.34 & 70.18 & 91.32 & 92.23 & \textbf{92.47} & 91.57 \\
    BelebeleRetrieval & test & 17.04 & 62.98 & 68.77 & \textbf{92.24} & 91.50 & 91.16 & 90.81 \\
    CmedqaRetrieval & dev & 1.81 & 5.09 & 7.42 & \textbf{35.89} & 32.48 & 31.98 & 31.73 \\
    CovidRetrieval & dev & 1.37 & 21.33 & 23.15 & \textbf{78.91} & 66.57 & 68.60 & 72.18 \\
    DuRetrieval & dev & 3.92 & 19.05 & 27.65 & \textbf{83.11} & 76.88 & 77.85 & 78.50 \\
    EcomRetrieval & dev & 4.33 & 13.65 & 22.54 & \textbf{60.68} & 29.77 & 47.77 & 55.86 \\
    LeCaRDv2 & test & 23.50 & 21.92 & 30.11 & \textbf{58.40} & 55.44 & 54.00 & 50.64 \\
    \makecell{MIRACLRetrieval \\ HardNegatives} & dev & 0.34 & 16.15 & 18.26 & \textbf{57.66} & 30.84 & 40.22 & 55.62 \\
    MLQARetrieval & test & 6.29 & 31.77 & 36.78 & \textbf{64.72} & 58.95 & 59.40 & 64.69 \\
    MMarcoRetrieval & dev & 7.06 & 28.30 & 37.94 & \textbf{79.66} & 70.05 & 69.96 & 76.28 \\
    MedicalRetrieval & dev & 1.32 & 7.73 & 11.96 & \textbf{56.63} & 52.92 & 51.84 & 51.15 \\
    \makecell{MultiLongDoc \\ Retrieval} & test & 0.54 & 3.32 & 3.51 & \textbf{17.17} & 10.79 & 11.37 & 7.78 \\
    \makecell{NeuCLIR2022Retrieval \\ HardNegatives} & test & 1.99 & 24.86 & 29.41 & \textbf{54.55} & 39.14 & 44.28 & 49.03 \\
    \makecell{NeuCLIR2023Retrieval \\ HardNegatives} & test & 2.50 & 24.44 & 30.14 & \textbf{50.10} & 42.90 & 43.89 & 49.84 \\
    PublicHealthQA & test & 17.66 & 59.55 & 65.23 & 84.56 & 85.38 & \textbf{86.30} & 84.94 \\
    T2Retrieval & dev & 2.77 & 21.45 & 28.94 & \textbf{83.16} & 72.88 & 76.19 & 77.76 \\
    VideoRetrieval & dev & 6.14 & 17.77 & 24.01 & \textbf{70.45} & 30.10 & 42.45 & 59.71 \\
    XPQARetrieval & test & 18.03 & 30.56 & 42.41 & \textbf{69.54} & 66.79 & 66.23 & 65.72 \\
    XQuADRetrieval & validation & 28.53 & 69.93 & 74.71 & 93.83 & 92.34 & 92.77 & \textbf{94.08} \\
    \makecell{mFollowIR \\ InstructionRetrieval} & test & 0.61 & -2.44 & \textbf{6.03} & 2.80 & 4.33 & 4.39 & 1.57 \\
    \midrule
    \multicolumn{2}{l|}{\textbf{Average}} & 8.09 & 27.04 & 32.96 & \textbf{64.27} & 55.11 & 57.66 & 60.47 \\
    \midrule
    \midrule
    \multicolumn{9}{c}{\textbf{STS} Spearman correlation on cosine similarity} \\
    \midrule
    AFQMC & validation & 7.65 & 10.97 & 14.60 & \textbf{38.85} & 35.27 & 36.19 & 36.56 \\
    ATEC & test & 14.16 & 13.99 & 21.44 & \textbf{44.80} & 42.25 & 42.87 & 43.59 \\
    BQ & test & 22.52 & 28.15 & 33.65 & 47.27 & 46.78 & 45.86 & \textbf{55.03} \\
    LCQMC & test & 20.96 & 40.26 & 52.80 & 74.47 & 73.56 & 74.16 & \textbf{75.42} \\
    PAWSX & test & 8.29 & 10.63 & 13.00 & 15.26 & 14.20 & \textbf{15.96} & 15.51 \\
    QBQTC & test & 17.95 & 22.54 & 22.54 & \textbf{34.35} & 30.09 & 32.60 & 32.74 \\
    STS22.v2 & test & 44.52 & 43.00 & 50.37 & \textbf{72.65} & 71.58 & 71.00 & 71.38 \\
    STSB & test & 34.50 & 65.25 & 72.05 & 81.35 & 80.18 & 80.52 & \textbf{82.70} \\
    \makecell{STSBenchmark \\ MultilingualSTS} & test & 36.19 & 65.46 & 71.87 & 82.42 & 80.78 & 81.15 & \textbf{82.75} \\
    \midrule
    \multicolumn{2}{l|}{\textbf{Average}} & 22.97 & 33.36 & 39.15 & 54.60 & 52.74 & 53.37 & \textbf{55.08} \\
    \midrule
    \bottomrule
    \end{tabular}
    \\
    \vspace{0.1in}
    \footnotesize{Evaluated on all tasks included in MTEB version $\mathrm{1.34.7}$, except the following: \texttt{MrTidyRetrieval}, \texttt{BrightRetrieval}, \texttt{MSMARCOv2}, \texttt{NeuCLIR2022Retrieval}, \texttt{NeuCLIR2023Retrieval} and \texttt{MIRACLRetrieval}. These tasks were excluded either due to bugs in the evaluation code or excessive computation times.} \\
  }
\end{table*}

\begin{table*}[ht]
  \centering
  \caption{Retrieval and STS evaluation on Hindi MTEB \citep{mteb} (1/1)}
  \label{appendix:results-mteb-hi-1}
  \setlength{\tabcolsep}{4.5pt}
  \vskip 0.1in
  \tiny{
    \begin{tabular}{cc|ccccccc}
    \toprule
    Task & Split & \texttt{{jc-v1}} & \texttt{{nllb-b}} & \texttt{{nllb-l}} & \texttt{{je-v3}} & \texttt{{jc-v2-s1}} & \texttt{{jc-v2-s2}} & \texttt{{jc-v2}} \\
    \midrule
    \midrule
    \multicolumn{9}{c}{\textbf{Retrieval} nDCG@10 [\%]} \\
    \midrule
    BelebeleRetrieval & test & 3.61 & 63.82 & 69.57 & \textbf{89.21} & 86.52 & 87.21 & 86.21 \\
    BelebeleRetrieval & test & 43.80 & 28.07 & 31.67 & \textbf{66.80} & 66.14 & 64.76 & 62.00 \\
    IndicQARetrieval & test & 2.11 & 32.46 & 35.63 & \textbf{67.26} & 63.22 & 62.60 & 63.87 \\
    \makecell{MIRACLRetrieval \\ HardNegatives} & dev & 0.14 & 21.87 & 23.79 & \textbf{56.47} & 43.20 & 43.77 & 54.80 \\
    MLQARetrieval & test & 1.82 & 34.81 & 40.69 & \textbf{63.77} & 59.08 & 58.41 & 62.40 \\
    MintakaRetrieval & test & 1.89 & 19.62 & 23.41 & 24.42 & 23.87 & 24.72 & \textbf{26.48} \\
    \makecell{MultiLongDoc \\ Retrieval} & test & 1.49 & 8.68 & 10.52 & 25.40 & 26.93 & \textbf{30.47} & 23.68 \\
    \makecell{WikipediaRetrieval \\ Multilingual} & test & 3.82 & 54.61 & 57.96 & \textbf{86.57} & 83.89 & 82.75 & 85.09 \\
    XPQARetrieval & test & 20.45 & 57.18 & 64.98 & \textbf{78.26} & 77.26 & 77.50 & 74.36 \\
    XQuADRetrieval & validation & 8.21 & 70.70 & 74.00 & \textbf{93.66} & 90.16 & 89.68 & 91.90 \\
    \midrule
    \multicolumn{2}{l|}{\textbf{Average}} & 8.73 & 39.18 & 43.22 & \textbf{65.18} & 62.03 & 62.19 & 63.08 \\
    \midrule
    \midrule
    \multicolumn{9}{c}{\textbf{STS} Spearman correlation on cosine similarity} \\
    \midrule
    SemRel24STS & test & 36.09 & 68.11 & 73.78 & 83.62 & 83.35 & \textbf{84.30} & 79.55 \\
    \midrule
    \multicolumn{2}{l|}{\textbf{Average}} & 36.09 & 68.11 & 73.78 & 83.62 & 83.35 & \textbf{84.30} & 79.55 \\
    \midrule
    \bottomrule
    \end{tabular}
    \\
    \vspace{0.1in}
    \footnotesize{Evaluated on all tasks included in MTEB version $\mathrm{1.34.7}$, except the following: \texttt{MrTidyRetrieval}, \texttt{BrightRetrieval}, \texttt{MSMARCOv2}, \texttt{NeuCLIR2022Retrieval}, \texttt{NeuCLIR2023Retrieval} and \texttt{MIRACLRetrieval}. These tasks were excluded either due to bugs in the evaluation code or excessive computation times.} \\
  }
\end{table*}

\begin{table*}[ht]
  \centering
  \caption{Retrieval and STS evaluation on German MTEB \citep{mteb} (1/1)}
  \label{appendix:results-mteb-de-1}
  \setlength{\tabcolsep}{4.5pt}
  \vskip 0.1in
  \tiny{
    \begin{tabular}{cc|ccccccc}
    \toprule
    Task & Split & \texttt{{jc-v1}} & \texttt{{nllb-b}} & \texttt{{nllb-l}} & \texttt{{je-v3}} & \texttt{{jc-v2-s1}} & \texttt{{jc-v2-s2}} & \texttt{{jc-v2}} \\
    \midrule
    \midrule
    \multicolumn{9}{c}{\textbf{Retrieval} nDCG@10 [\%]} \\
    \midrule
    BelebeleRetrieval & test & 54.13 & 39.84 & 59.47 & 92.99 & 93.20 & \textbf{93.21} & 92.08 \\
    GerDaLIR & test & 2.14 & 0.14 & 0.41 & \textbf{16.18} & 9.06 & 10.57 & 2.22 \\
    GerDaLIRSmall & test & 6.18 & 0.47 & 1.34 & \textbf{36.66} & 21.23 & 25.33 & 6.76 \\
    GermanDPR & test & 49.26 & 35.74 & 45.63 & \textbf{82.47} & 80.26 & 81.00 & 80.52 \\
    \makecell{GermanGovService \\ Retrieval} & test & 45.22 & 37.40 & 47.70 & \textbf{88.71} & 83.07 & 82.75 & 87.18 \\
    GermanQuAD-Retrieval & test & 62.67 & 47.35 & 61.08 & \textbf{94.15} & 90.77 & 91.66 & 92.80 \\
    LegalQuAD & test & 14.97 & 4.06 & 11.18 & \textbf{58.83} & 41.33 & 46.25 & 32.87 \\
    \makecell{MIRACLRetrieval \\ HardNegatives} & dev & 19.37 & 13.07 & 20.59 & \textbf{53.37} & 39.41 & 42.67 & 51.61 \\
    MLQARetrieval & test & 33.09 & 17.87 & 28.39 & \textbf{67.50} & 63.06 & 61.94 & 65.77 \\
    MintakaRetrieval & test & 20.37 & 9.66 & 17.62 & 27.05 & 29.08 & 30.10 & \textbf{31.55} \\
    \makecell{MultiLongDoc \\ Retrieval} & test & 13.25 & 3.15 & 9.74 & 37.81 & 34.15 & \textbf{38.18} & 24.10 \\
    \makecell{WikipediaRetrieval \\ Multilingual} & test & 62.29 & 44.00 & 58.90 & \textbf{89.34} & 89.01 & 88.04 & 88.75 \\
    XMarket & test & 8.78 & 4.86 & 7.76 & \textbf{28.28} & 12.12 & 18.33 & 16.54 \\
    XPQARetrieval & test & 52.72 & 32.65 & 48.28 & \textbf{84.68} & 83.44 & 83.50 & 82.31 \\
    XQuADRetrieval & validation & 62.08 & 49.82 & 66.27 & \textbf{95.79} & 94.18 & 93.79 & 94.90 \\
    \midrule
    \multicolumn{2}{l|}{\textbf{Average}} & 33.77 & 22.67 & 32.29 & \textbf{63.59} & 57.56 & 59.16 & 56.66 \\
    \midrule
    \midrule
    \multicolumn{9}{c}{\textbf{STS} Spearman correlation on cosine similarity} \\
    \midrule
    GermanSTSBenchmark & test & 63.94 & 54.53 & 66.52 & 82.98 & 82.17 & 82.25 & \textbf{84.82} \\
    STS22.v2 & test & 22.42 & 10.39 & 27.08 & 59.16 & 59.22 & \textbf{61.35} & 58.30 \\
    \makecell{STSBenchmark \\ MultilingualSTS} & test & 63.78 & 53.82 & 66.77 & 83.83 & 83.14 & 83.33 & \textbf{85.68} \\
    \midrule
    \multicolumn{2}{l|}{\textbf{Average}} & 50.04 & 39.58 & 53.45 & 75.32 & 74.84 & 75.64 & \textbf{76.27} \\
    \midrule
    \bottomrule
    \end{tabular}
    \\
    \vspace{0.1in}
    \footnotesize{Evaluated on all tasks included in MTEB version $\mathrm{1.34.7}$, except the following: \texttt{MrTidyRetrieval}, \texttt{BrightRetrieval}, \texttt{MSMARCOv2}, \texttt{NeuCLIR2022Retrieval}, \texttt{NeuCLIR2023Retrieval} and \texttt{MIRACLRetrieval}. These tasks were excluded either due to bugs in the evaluation code or excessive computation times.} \\
  }
\end{table*}

\begin{table*}[ht]
  \centering
  \caption{Retrieval and STS evaluation on French MTEB \citep{mteb} (1/1)}
  \label{appendix:results-mteb-fr-1}
  \setlength{\tabcolsep}{4.5pt}
  \vskip 0.1in
  \tiny{
    \begin{tabular}{cc|ccccccc}
    \toprule
    Task & Split & \texttt{{jc-v1}} & \texttt{{nllb-b}} & \texttt{{nllb-l}} & \texttt{{je-v3}} & \texttt{{jc-v2-s1}} & \texttt{{jc-v2-s2}} & \texttt{{jc-v2}} \\
    \midrule
    \midrule
    \multicolumn{9}{c}{\textbf{Retrieval} nDCG@10 [\%]} \\
    \midrule
    AlloprofRetrieval & test & 22.90 & 12.36 & 18.70 & \textbf{54.39} & 47.52 & 51.75 & 51.24 \\
    BSARDRetrieval & test & 25.68 & 19.37 & 36.49 & 63.96 & 64.86 & \textbf{65.77} & 58.11 \\
    BelebeleRetrieval & test & 66.86 & 53.05 & 71.63 & \textbf{93.86} & 90.26 & 90.65 & 92.38 \\
    FQuADRetrieval & test & 51.07 & 36.70 & 43.68 & \textbf{74.33} & 68.04 & 67.03 & 70.59 \\
    \makecell{MIRACLRetrieval \\ HardNegatives} & dev & 26.57 & 17.58 & 19.93 & \textbf{55.22} & 38.43 & 42.81 & 52.72 \\
    MintakaRetrieval & test & 22.91 & 15.57 & 20.15 & 26.94 & 28.39 & 29.17 & \textbf{30.79} \\
    \makecell{MultiLongDoc \\ Retrieval} & test & 41.80 & 20.64 & 26.17 & 59.84 & 59.76 & \textbf{66.51} & 54.17 \\
    PublicHealthQA & test & 64.54 & 51.62 & 64.67 & \textbf{92.05} & 88.13 & 90.71 & 91.75 \\
    \makecell{StatcanDialogue \\ DatasetRetrieval} & test & 2.31 & 5.09 & 11.42 & \textbf{22.68} & 4.63 & 10.65 & 17.13 \\
    SyntecRetrieval & test & 64.17 & 44.47 & 63.52 & \textbf{84.03} & 77.33 & 78.95 & 79.01 \\
    XPQARetrieval & test & 51.10 & 34.80 & 49.81 & \textbf{77.68} & 76.80 & 77.56 & 74.84 \\
    \midrule
    \multicolumn{2}{l|}{\textbf{Average}} & 39.99 & 28.30 & 38.74 & \textbf{64.09} & 58.56 & 61.05 & 61.16 \\
    \midrule
    \midrule
    \multicolumn{9}{c}{\textbf{STS} Spearman correlation on cosine similarity} \\
    \midrule
    SICKFr & test & 67.42 & 63.22 & 68.86 & 76.51 & 77.38 & 77.02 & \textbf{80.55} \\
    STS22.v2 & test & 66.65 & 61.01 & 73.31 & \textbf{83.48} & 77.99 & 81.16 & 79.90 \\
    \makecell{STSBenchmark \\ MultilingualSTS} & test & 69.82 & 67.60 & 74.97 & 83.33 & 81.79 & 82.40 & \textbf{84.65} \\
    \midrule
    \multicolumn{2}{l|}{\textbf{Average}} & 67.97 & 63.94 & 72.38 & 81.11 & 79.06 & 80.19 & \textbf{81.70} \\
    \midrule
    \bottomrule
    \end{tabular}
    \\
    \vspace{0.1in}
    \footnotesize{Evaluated on all tasks included in MTEB version $\mathrm{1.34.7}$, except the following: \texttt{MrTidyRetrieval}, \texttt{BrightRetrieval}, \texttt{MSMARCOv2}, \texttt{NeuCLIR2022Retrieval}, \texttt{NeuCLIR2023Retrieval} and \texttt{MIRACLRetrieval}. These tasks were excluded either due to bugs in the evaluation code or excessive computation times.} \\
  }
\end{table*}

\begin{table*}[ht]
  \centering
  \caption{Retrieval and STS evaluation on Spanish MTEB \citep{mteb} (1/1)}
  \label{appendix:results-mteb-es-1}
  \setlength{\tabcolsep}{4.5pt}
  \vskip 0.1in
  \tiny{
    \begin{tabular}{cc|ccccccc}
    \toprule
    Task & Split & \texttt{{jc-v1}} & \texttt{{nllb-b}} & \texttt{{nllb-l}} & \texttt{{je-v3}} & \texttt{{jc-v2-s1}} & \texttt{{jc-v2-s2}} & \texttt{{jc-v2}} \\
    \midrule
    \midrule
    \multicolumn{9}{c}{\textbf{Retrieval} nDCG@10 [\%]} \\
    \midrule
    BelebeleRetrieval & test & 62.63 & 52.86 & 69.01 & \textbf{93.52} & 91.81 & 91.42 & 92.36 \\
    \makecell{MIRACLRetrieval \\ HardNegatives} & dev & 27.05 & 16.71 & 22.04 & \textbf{51.08} & 43.17 & 45.33 & 50.94 \\
    MLQARetrieval & test & 42.28 & 29.25 & 40.56 & \textbf{68.62} & 62.89 & 61.50 & 66.69 \\
    MintakaRetrieval & test & 21.47 & 16.29 & 20.21 & 26.93 & 29.00 & 29.76 & \textbf{31.28} \\
    \makecell{MultiLongDoc \\ Retrieval} & test & 41.13 & 14.04 & 20.55 & 62.07 & 63.86 & \textbf{67.68} & 55.56 \\
    PublicHealthQA & test & 53.89 & 46.09 & 61.15 & 83.09 & 83.61 & \textbf{86.62} & 83.75 \\
    \makecell{SpanishPassage \\ RetrievalS2P} & test & 20.02 & 18.03 & 28.32 & \textbf{43.11} & 35.95 & 38.94 & 41.72 \\
    \makecell{SpanishPassage \\ RetrievalS2S} & test & 38.77 & 37.06 & 48.75 & 69.71 & 70.12 & 70.79 & \textbf{72.98} \\
    XMarket & test & 12.02 & 6.38 & 10.22 & \textbf{26.69} & 14.20 & 20.96 & 17.51 \\
    XPQARetrieval & test & 44.80 & 29.19 & 42.81 & \textbf{72.14} & 71.04 & 71.36 & 69.10 \\
    XQuADRetrieval & validation & 73.39 & 64.76 & 75.47 & \textbf{95.67} & 92.84 & 92.20 & 94.40 \\
    \midrule
    \multicolumn{2}{l|}{\textbf{Average}} & 39.77 & 30.06 & 39.92 & \textbf{62.97} & 59.86 & 61.51 & 61.48 \\
    \midrule
    \midrule
    \multicolumn{9}{c}{\textbf{STS} Spearman correlation on cosine similarity} \\
    \midrule
    STS17 & test & 77.09 & 77.23 & 83.26 & 86.93 & \textbf{87.31} & 86.99 & 87.29 \\
    STS22.v2 & test & 57.19 & 46.44 & 53.42 & 75.58 & 75.16 & \textbf{76.48} & 76.01 \\
    \makecell{STSBenchmark \\ MultilingualSTS} & test & 71.90 & 65.44 & 75.15 & 84.72 & 82.76 & 83.35 & \textbf{85.90} \\
    STSES & test & 74.00 & 72.02 & 66.90 & 75.89 & 76.47 & 75.22 & \textbf{78.34} \\
    \midrule
    \multicolumn{2}{l|}{\textbf{Average}} & 70.05 & 65.28 & 69.68 & 80.78 & 80.42 & 80.51 & \textbf{81.89} \\
    \midrule
    \bottomrule
    \end{tabular}
    \\
    \vspace{0.1in}
    \footnotesize{Evaluated on all tasks included in MTEB version $\mathrm{1.34.7}$, except the following: \texttt{MrTidyRetrieval}, \texttt{BrightRetrieval}, \texttt{MSMARCOv2}, \texttt{NeuCLIR2022Retrieval}, \texttt{NeuCLIR2023Retrieval} and \texttt{MIRACLRetrieval}. These tasks were excluded either due to bugs in the evaluation code or excessive computation times.} \\
  }
\end{table*}

\begin{table*}[ht]
  \centering
  \caption{Retrieval and STS evaluation on Japanese MTEB \citep{mteb} (1/1)}
  \label{appendix:results-mteb-jp-1}
  \setlength{\tabcolsep}{4.5pt}
  \vskip 0.1in
  \tiny{
    \begin{tabular}{cc|ccccccc}
    \toprule
    Task & Split & \texttt{{jc-v1}} & \texttt{{nllb-b}} & \texttt{{nllb-l}} & \texttt{{je-v3}} & \texttt{{jc-v2-s1}} & \texttt{{jc-v2-s2}} & \texttt{{jc-v2}} \\
    \midrule
    \midrule
    \multicolumn{9}{c}{\textbf{Retrieval} nDCG@10 [\%]} \\
    \midrule
    BelebeleRetrieval & test & 17.69 & 64.01 & 69.81 & \textbf{91.73} & 91.12 & 90.84 & 90.23 \\
    JaGovFaqsRetrieval & test & 8.85 & 34.49 & 40.39 & 71.92 & \textbf{72.73} & 72.10 & 70.31 \\
    JaQuADRetrieval & validation & 8.62 & 29.17 & 32.74 & \textbf{54.78} & 51.98 & 53.23 & 52.43 \\
    JaqketRetrieval & test & 0.09 & 19.51 & 26.41 & \textbf{46.91} & 25.71 & 26.54 & 32.70 \\
    \makecell{MIRACLRetrieval \\ HardNegatives} & dev & 0.82 & 17.52 & 18.26 & \textbf{66.26} & 47.37 & 50.65 & 57.92 \\
    MintakaRetrieval & test & 5.38 & 15.54 & 17.89 & 20.68 & 24.43 & 24.77 & \textbf{25.57} \\
    \makecell{MultiLongDoc \\ Retrieval} & test & 0.49 & 8.79 & 10.37 & 38.40 & 28.15 & \textbf{38.88} & 33.15 \\
    \makecell{NLPJournal \\ AbsIntroRetrieval} & test & 22.23 & 36.99 & 37.98 & \textbf{98.75} & 96.25 & 97.28 & 97.78 \\
    \makecell{NLPJournal \\ TitleAbsRetrieval} & test & 17.13 & 49.47 & 56.92 & 94.73 & 95.53 & 95.25 & \textbf{95.77} \\
    \makecell{NLPJournal \\ TitleIntroRetrieval} & test & 4.95 & 19.20 & 23.94 & \textbf{93.40} & 87.58 & 90.32 & 91.21 \\
    XPQARetrieval & test & 32.25 & 43.38 & 56.12 & 75.55 & 75.89 & \textbf{76.56} & 73.22 \\
    \midrule
    \multicolumn{2}{l|}{\textbf{Average}} & 10.77 & 30.73 & 35.53 & \textbf{68.47} & 63.34 & 65.13 & 65.48 \\
    \midrule
    \midrule
    \multicolumn{9}{c}{\textbf{STS} Spearman correlation on cosine similarity} \\
    \midrule
    JSICK & test & 60.57 & 77.93 & 78.38 & \textbf{81.38} & 80.56 & 80.28 & 80.45 \\
    JSTS & validation & 51.09 & 76.85 & 80.11 & 80.69 & 82.39 & 82.64 & \textbf{83.57} \\
    \midrule
    \multicolumn{2}{l|}{\textbf{Average}} & 55.83 & 77.39 & 79.25 & 81.03 & 81.48 & 81.46 & \textbf{82.01} \\
    \midrule
    \bottomrule
    \end{tabular}
    \\
    \vspace{0.1in}
    \footnotesize{Evaluated on all tasks included in MTEB version $\mathrm{1.34.7}$, except the following: \texttt{MrTidyRetrieval}, \texttt{BrightRetrieval}, \texttt{MSMARCOv2}, \texttt{NeuCLIR2022Retrieval}, \texttt{NeuCLIR2023Retrieval} and \texttt{MIRACLRetrieval}. These tasks were excluded either due to bugs in the evaluation code or excessive computation times.} \\
  }
\end{table*}

\begin{table*}[ht]
  \centering
  \caption{Retrieval and STS evaluation on Russian MTEB \citep{mteb} (1/1)}
  \label{appendix:results-mteb-ru-1}
  \setlength{\tabcolsep}{4.5pt}
  \vskip 0.1in
  \tiny{
    \begin{tabular}{cc|ccccccc}
    \toprule
    Task & Split & \texttt{{jc-v1}} & \texttt{{nllb-b}} & \texttt{{nllb-l}} & \texttt{{je-v3}} & \texttt{{jc-v2-s1}} & \texttt{{jc-v2-s2}} & \texttt{{jc-v2}} \\
    \midrule
    \midrule
    \multicolumn{9}{c}{\textbf{Retrieval} nDCG@10 [\%]} \\
    \midrule
    BelebeleRetrieval & test & 9.30 & 63.01 & 71.61 & \textbf{93.11} & 90.74 & 90.96 & 92.41 \\
    \makecell{MIRACLRetrieval \\ HardNegatives} & dev & 1.15 & 18.44 & 21.74 & \textbf{65.29} & 42.54 & 49.71 & 59.74 \\
    \makecell{MultiLongDoc \\ Retrieval} & test & 2.85 & 15.13 & 20.25 & 49.54 & 54.01 & \textbf{58.75} & 46.55 \\
    \makecell{NeuCLIR2022Retrieval \\ HardNegatives} & test & 0.97 & 29.00 & 36.59 & \textbf{58.08} & 47.32 & 47.36 & 52.48 \\
    \makecell{NeuCLIR2023Retrieval \\ HardNegatives} & test & 4.26 & 28.95 & 34.23 & \textbf{52.91} & 47.55 & 49.54 & 51.30 \\
    PublicHealthQA & test & 16.70 & 55.78 & 60.23 & 84.49 & 85.86 & \textbf{87.28} & 84.79 \\
    RiaNewsRetrieval & test & 0.86 & 26.68 & 37.02 & 79.17 & 73.18 & 79.61 & \textbf{81.43} \\
    \makecell{RiaNewsRetrieval \\ HardNegatives} & test & 0.78 & 30.50 & 39.94 & 81.09 & 74.08 & 80.77 & \textbf{82.99} \\
    RuBQRetrieval & test & 2.54 & 27.09 & 29.74 & \textbf{72.27} & 60.34 & 62.11 & 69.12 \\
    XQuADRetrieval & validation & 14.28 & 69.50 & 75.24 & \textbf{95.26} & 91.97 & 91.53 & 94.15 \\
    \makecell{mFollowIR \\ InstructionRetrieval} & test & -1.07 & -4.64 & 0.97 & -0.20 & 0.60 & \textbf{2.66} & -2.92 \\
    \midrule
    \multicolumn{2}{l|}{\textbf{Average}} & 4.78 & 32.68 & 38.87 & \textbf{66.46} & 60.74 & 63.66 & 64.73 \\
    \midrule
    \midrule
    \multicolumn{9}{c}{\textbf{STS} Spearman correlation on cosine similarity} \\
    \midrule
    RUParaPhraserSTS & test & 40.46 & 51.46 & 49.95 & \textbf{74.57} & 69.62 & 71.27 & 74.14 \\
    RuSTSBenchmarkSTS & test & 51.02 & 70.40 & 76.15 & 81.54 & 81.52 & 81.36 & \textbf{83.67} \\
    STS22.v2 & test & 16.95 & 38.21 & 48.97 & \textbf{71.00} & 69.81 & 70.75 & 69.92 \\
    \makecell{STSBenchmark \\ MultilingualSTS} & test & 50.63 & 70.13 & 76.28 & 81.53 & 81.53 & 81.36 & \textbf{83.44} \\
    \midrule
    \multicolumn{2}{l|}{\textbf{Average}} & 39.77 & 57.55 & 62.84 & 77.16 & 75.62 & 76.18 & \textbf{77.79} \\
    \midrule
    \bottomrule
    \end{tabular}
    \\
    \vspace{0.1in}
    \footnotesize{Evaluated on all tasks included in MTEB version $\mathrm{1.34.7}$, except the following: \texttt{MrTidyRetrieval}, \texttt{BrightRetrieval}, \texttt{MSMARCOv2}, \texttt{NeuCLIR2022Retrieval}, \texttt{NeuCLIR2023Retrieval} and \texttt{MIRACLRetrieval}. These tasks were excluded either due to bugs in the evaluation code or excessive computation times.} \\
  }
\end{table*}

\begin{table*}[ht]
  \centering
  \caption{Retrieval and STS evaluation on Polish MTEB \citep{mteb} (1/1)}
  \label{appendix:results-mteb-pl-1}
  \setlength{\tabcolsep}{4.5pt}
  \vskip 0.1in
  \tiny{
    \begin{tabular}{cc|ccccccc}
    \toprule
    Task & Split & \texttt{{jc-v1}} & \texttt{{nllb-b}} & \texttt{{nllb-l}} & \texttt{{je-v3}} & \texttt{{jc-v2-s1}} & \texttt{{jc-v2-s2}} & \texttt{{jc-v2}} \\
    \midrule
    \midrule
    \multicolumn{9}{c}{\textbf{Retrieval} nDCG@10 [\%]} \\
    \midrule
    ArguAna-PL & test & 7.46 & 10.03 & 22.07 & 38.52 & 43.92 & \textbf{46.80} & 40.45 \\
    BelebeleRetrieval & test & 30.29 & 35.21 & 53.19 & \textbf{92.45} & 89.16 & 89.63 & 90.97 \\
    DBPedia-PL & test & 13.01 & 9.41 & 13.34 & \textbf{34.88} & 24.07 & 24.83 & 30.45 \\
    \makecell{DBPedia-PL \\ HardNegatives} & test & 16.36 & 15.28 & 18.87 & \textbf{38.80} & 27.67 & 27.98 & 35.09 \\
    FiQA-PL & test & 2.47 & 3.10 & 6.56 & \textbf{38.85} & 26.67 & 29.02 & 32.71 \\
    HotpotQA-PL & test & 19.09 & 12.95 & 16.80 & \textbf{60.29} & 49.51 & 48.40 & 55.11 \\
    \makecell{HotpotQA-PL \\ HardNegatives} & test & 24.86 & 19.64 & 23.17 & \textbf{62.17} & 51.89 & 51.08 & 56.15 \\
    MSMARCO-PL & test & 11.30 & 7.83 & 16.01 & \textbf{64.78} & 43.29 & 49.89 & 59.49 \\
    \makecell{MSMARCO-PL \\ HardNegatives} & test & 32.40 & 22.79 & 28.04 & \textbf{66.86} & 52.27 & 55.61 & 63.51 \\
    NFCorpus-PL & test & 7.40 & 9.99 & 11.26 & \textbf{31.39} & 27.00 & 28.71 & 27.66 \\
    NQ-PL & test & 3.25 & 4.29 & 7.43 & \textbf{54.01} & 29.31 & 30.78 & 43.29 \\
    NQ-PLHardNegatives & test & 5.27 & 6.57 & 10.43 & \textbf{56.20} & 32.12 & 32.49 & 45.98 \\
    Quora-PL & test & 52.30 & 48.28 & 58.88 & 53.57 & 77.74 & 79.20 & \textbf{80.65} \\
    \makecell{Quora-PL \\ HardNegatives} & test & 54.25 & 49.15 & 60.60 & 54.82 & 77.87 & 79.01 & \textbf{80.57} \\
    SCIDOCS-PL & test & 3.14 & 4.38 & 6.60 & \textbf{15.38} & 11.96 & 14.35 & 13.73 \\
    SciFact-PL & test & 8.82 & 18.42 & 23.21 & \textbf{64.88} & 61.69 & 60.32 & 57.68 \\
    TRECCOVID-PL & test & 12.15 & 16.74 & 30.85 & \textbf{71.73} & 51.25 & 58.40 & 70.81 \\
    XPQARetrieval & test & 23.76 & 15.67 & 26.19 & 53.77 & 54.00 & \textbf{54.60} & 52.33 \\
    \midrule
    \multicolumn{2}{l|}{\textbf{Average}} & 18.20 & 17.21 & 24.08 & \textbf{52.96} & 46.19 & 47.84 & 52.04 \\
    \midrule
    \midrule
    \multicolumn{9}{c}{\textbf{STS} Spearman correlation on cosine similarity} \\
    \midrule
    CDSC-R & test & 78.79 & 76.12 & 85.32 & \textbf{91.97} & 90.95 & 91.20 & 91.16 \\
    SICK-R-PL & test & 53.67 & 48.13 & 53.26 & 72.90 & 74.85 & 75.11 & \textbf{78.52} \\
    STS22.v2 & test & 20.31 & 16.65 & 30.84 & 49.36 & 45.12 & \textbf{50.53} & 45.73 \\
    \makecell{STSBenchmark \\ MultilingualSTS} & test & 57.05 & 49.94 & 61.56 & 82.05 & 80.90 & 81.04 & \textbf{82.62} \\
    \midrule
    \multicolumn{2}{l|}{\textbf{Average}} & 52.46 & 47.71 & 57.75 & 74.07 & 72.95 & 74.47 & \textbf{74.51} \\
    \midrule
    \bottomrule
    \end{tabular}
    \\
    \vspace{0.1in}
    \footnotesize{Evaluated on all tasks included in MTEB version $\mathrm{1.34.7}$, except the following: \texttt{MrTidyRetrieval}, \texttt{BrightRetrieval}, \texttt{MSMARCOv2}, \texttt{NeuCLIR2022Retrieval}, \texttt{NeuCLIR2023Retrieval} and \texttt{MIRACLRetrieval}. These tasks were excluded either due to bugs in the evaluation code or excessive computation times.} \\
  }
\end{table*}

\begin{table*}[ht]
  \centering
  \caption{Retrieval and STS evaluation on Cross-lingual MTEB \citep{mteb} (1/2)}
  \label{appendix:results-mteb-xl-1}
  \setlength{\tabcolsep}{4.5pt}
  \vskip 0.1in
  \tiny{
    \begin{tabular}{ccc|ccccccc}
    \toprule
    Task & Split & Languages & \texttt{{jc-v1}} & \texttt{{nllb-b}} & \texttt{{nllb-l}} & \texttt{{je-v3}} & \texttt{{jc-v2-s1}} & \texttt{{jc-v2-s2}} & \texttt{{jc-v2}} \\
    \midrule
    \midrule
    \multicolumn{10}{c}{\textbf{Retrieval} nDCG@10 [\%]} \\
    \midrule
    BelebeleRetrieval & test & de,en & 66.96 & 52.96 & 69.05 & 92.38 & 92.11 & 92.36 & \textbf{92.40} \\
    BelebeleRetrieval & test & en,de & 46.64 & 35.59 & 62.72 & \textbf{91.80} & 90.18 & 91.19 & 89.84 \\
    BelebeleRetrieval & test & fr,en & 79.80 & 64.85 & 73.07 & \textbf{92.75} & 88.64 & 90.50 & 92.32 \\
    BelebeleRetrieval & test & en,fr & 62.43 & 54.01 & 74.08 & \textbf{92.16} & 89.35 & 91.16 & 90.53 \\
    BelebeleRetrieval & test & hi,en & 8.07 & 68.55 & 71.82 & \textbf{91.48} & 87.95 & 89.73 & 90.46 \\
    BelebeleRetrieval & test & en,hi & 2.88 & 65.36 & 72.27 & \textbf{87.67} & 82.39 & 85.74 & 83.21 \\
    BelebeleRetrieval & test & hi,en & 55.56 & 33.15 & 36.43 & \textbf{70.16} & 68.62 & 68.77 & 69.62 \\
    BelebeleRetrieval & test & en,hi & 44.96 & 24.18 & 32.34 & \textbf{57.53} & 51.68 & 54.83 & 51.97 \\
    BelebeleRetrieval & test & jp,en & 25.11 & 67.58 & 71.78 & 91.17 & 88.68 & 90.24 & \textbf{91.24} \\
    BelebeleRetrieval & test & en,jp & 7.18 & 60.99 & 70.07 & \textbf{89.49} & 86.80 & 88.90 & 85.39 \\
    BelebeleRetrieval & test & pl,en & 45.49 & 48.50 & 63.12 & \textbf{92.22} & 88.98 & 90.17 & 91.70 \\
    BelebeleRetrieval & test & en,pl & 28.36 & 36.52 & 58.30 & \textbf{90.63} & 89.00 & 90.34 & 88.36 \\
    BelebeleRetrieval & test & ru,en & 15.23 & 64.24 & 71.26 & \textbf{91.70} & 87.98 & 89.50 & 91.25 \\
    BelebeleRetrieval & test & en,ru & 7.67 & 63.76 & 73.50 & \textbf{91.82} & 89.75 & 91.20 & 90.16 \\
    BelebeleRetrieval & test & es,en & 74.95 & 64.49 & 72.92 & \textbf{92.65} & 90.23 & 90.84 & 91.79 \\
    BelebeleRetrieval & test & en,es & 58.15 & 53.79 & 72.11 & \textbf{91.33} & 89.96 & 91.11 & 90.96 \\
    BelebeleRetrieval & test & zh,en & 31.42 & 62.78 & 68.42 & 89.68 & 88.79 & 89.78 & \textbf{89.86} \\
    BelebeleRetrieval & test & en,zh & 8.69 & 61.78 & 69.85 & \textbf{89.43} & 85.59 & 88.81 & 86.22 \\
    BelebeleRetrieval & test & zh,en & 27.30 & 62.34 & 67.54 & \textbf{89.52} & 87.45 & 88.94 & 89.48 \\
    BelebeleRetrieval & test & en,zh & 7.50 & 61.05 & 69.84 & \textbf{89.92} & 86.40 & 88.74 & 85.50 \\
    BelebeleRetrieval & test & hi,hi & 6.41 & 22.82 & 31.82 & \textbf{58.73} & 52.19 & 57.18 & 51.35 \\
    BelebeleRetrieval & test & hi,hi & 2.23 & 31.65 & 35.49 & \textbf{67.89} & 65.01 & 65.83 & 64.99 \\
    CUREv1 & dentistry\_and\_oral\_health & es,en & 9.53 & 11.85 & 22.29 & 44.83 & 37.08 & 40.17 & \textbf{47.88} \\
    CUREv1 & dermatology & es,en & 21.74 & 12.79 & 18.25 & 50.08 & 36.11 & 41.21 & \textbf{56.62} \\
    CUREv1 & gastroenterology & es,en & 18.31 & 12.20 & 20.48 & 48.85 & 38.38 & 41.16 & \textbf{51.63} \\
    CUREv1 & genetics & es,en & 19.82 & 11.87 & 21.38 & \textbf{51.52} & 39.55 & 40.01 & 50.42 \\
    CUREv1 & neuroscience\_and\_neurology & es,en & 9.30 & 9.07 & 18.68 & \textbf{41.36} & 29.13 & 31.69 & 40.34 \\
    CUREv1 & orthopedic\_surgery & es,en & 8.69 & 5.04 & 15.17 & 37.66 & 39.86 & 38.10 & \textbf{43.00} \\
    CUREv1 & otorhinolaryngology & es,en & 8.30 & 6.23 & 14.92 & 38.43 & 32.59 & 33.32 & \textbf{44.43} \\
    CUREv1 & plastic\_surgery & es,en & 11.80 & 8.55 & 18.93 & 42.34 & 34.48 & 36.89 & \textbf{45.23} \\
    CUREv1 & psychiatry\_and\_psychology & es,en & 14.34 & 15.93 & 24.81 & 50.45 & 39.39 & 41.25 & \textbf{50.48} \\
    CUREv1 & pulmonology & es,en & 13.17 & 13.76 & 23.95 & \textbf{48.60} & 36.02 & 36.85 & 47.91 \\
    CUREv1 & avg & es,en & 13.50 & 10.73 & 19.89 & 45.41 & 36.26 & 38.06 & \textbf{47.79} \\
    CUREv1 & dentistry\_and\_oral\_health & fr,en & 12.04 & 12.18 & 21.30 & 42.74 & 34.24 & 37.53 & \textbf{48.03} \\
    CUREv1 & dermatology & fr,en & 27.76 & 15.22 & 19.91 & 50.74 & 34.70 & 38.40 & \textbf{56.08} \\
    CUREv1 & gastroenterology & fr,en & 21.87 & 12.85 & 19.89 & 48.14 & 34.75 & 37.14 & \textbf{50.20} \\
    CUREv1 & genetics & fr,en & 26.69 & 13.79 & 22.69 & \textbf{49.52} & 33.82 & 34.83 & 47.66 \\
    CUREv1 & neuroscience\_and\_neurology & fr,en & 14.39 & 10.59 & 16.83 & 38.74 & 24.63 & 27.98 & \textbf{39.30} \\
    CUREv1 & orthopedic\_surgery & fr,en & 14.62 & 6.92 & 12.95 & 35.61 & 35.70 & 36.68 & \textbf{44.31} \\
    CUREv1 & otorhinolaryngology & fr,en & 10.57 & 7.09 & 14.12 & 40.14 & 29.33 & 30.62 & \textbf{44.08} \\
    CUREv1 & plastic\_surgery & fr,en & 20.62 & 9.15 & 19.19 & 42.27 & 33.00 & 34.33 & \textbf{45.15} \\
    CUREv1 & psychiatry\_and\_psychology & fr,en & 20.73 & 14.97 & 25.78 & \textbf{49.75} & 36.05 & 37.79 & 48.87 \\
    CUREv1 & pulmonology & fr,en & 16.56 & 12.42 & 22.73 & \textbf{46.90} & 32.95 & 33.96 & 45.83 \\
    CUREv1 & avg & fr,en & 18.58 & 11.52 & 19.54 & 44.46 & 32.92 & 34.93 & \textbf{46.95} \\
    \makecell{CrossLingualSemantic \\ DiscriminationWMT19} & test & de,fr & 31.64 & 45.69 & 70.74 & 83.98 & 89.75 & 89.34 & \textbf{91.11} \\
    \makecell{CrossLingualSemantic \\ DiscriminationWMT19} & test & fr,de & 30.14 & 38.42 & 73.18 & 81.67 & 89.41 & 89.00 & \textbf{90.83} \\
    \makecell{CrossLingualSemantic \\ DiscriminationWMT21} & test & de,fr & 37.51 & 53.97 & 72.56 & 81.64 & \textbf{89.81} & 88.24 & 89.81 \\
    \makecell{CrossLingualSemantic \\ DiscriminationWMT21} & test & fr,de & 39.98 & 53.41 & 77.38 & 83.09 & 84.21 & 84.77 & \textbf{87.23} \\
    MLQARetrieval & test & de,en & 45.85 & 26.14 & 35.89 & \textbf{66.97} & 60.34 & 60.86 & 65.22 \\
    MLQARetrieval & test & de,es & 34.32 & 27.08 & 39.11 & \textbf{70.58} & 65.32 & 65.55 & 69.11 \\
    MLQARetrieval & test & de,hi & 2.54 & 31.57 & 42.82 & \textbf{68.43} & 62.34 & 63.67 & 65.91 \\
    MLQARetrieval & test & de,zh & 6.91 & 26.13 & 35.68 & \textbf{66.87} & 59.62 & 62.15 & 64.21 \\
    MLQARetrieval & test & en,de & 33.03 & 17.24 & 32.36 & \textbf{69.30} & 66.14 & 66.89 & 67.81 \\
    MLQARetrieval & test & en,es & 40.98 & 28.19 & 40.57 & \textbf{66.38} & 62.79 & 63.06 & 64.67 \\
    MLQARetrieval & test & en,hi & 2.10 & 35.97 & 43.03 & \textbf{61.04} & 57.01 & 58.68 & 59.43 \\
    MLQARetrieval & test & en,zh & 6.43 & 28.74 & 35.76 & \textbf{59.75} & 56.45 & 58.25 & 58.11 \\
    MLQARetrieval & test & es,de & 32.10 & 21.12 & 36.79 & \textbf{73.89} & 68.02 & 67.89 & 71.58 \\
    MLQARetrieval & test & es,en & 50.20 & 34.70 & 42.96 & \textbf{68.75} & 61.07 & 61.07 & 67.18 \\
    MLQARetrieval & test & es,hi & 2.38 & 40.29 & 49.28 & \textbf{68.74} & 62.01 & 62.47 & 66.13 \\
    MLQARetrieval & test & es,zh & 6.91 & 32.84 & 41.74 & \textbf{68.56} & 62.00 & 63.32 & 66.16 \\
    \bottomrule
    \end{tabular}
    \\
    \vspace{0.1in}
    \footnotesize{Evaluated on all tasks included in MTEB version $\mathrm{1.34.7}$, except the following: \texttt{MrTidyRetrieval}, \texttt{BrightRetrieval}, \texttt{MSMARCOv2}, \texttt{NeuCLIR2022Retrieval}, \texttt{NeuCLIR2023Retrieval} and \texttt{MIRACLRetrieval}. These tasks were excluded either due to bugs in the evaluation code or excessive computation times.} \\
  }
\end{table*}

\begin{table*}[ht]
  \centering
  \caption{Retrieval and STS evaluation on Cross-lingual MTEB \citep{mteb} (2/2)}
  \label{appendix:results-mteb-xl-2}
  \setlength{\tabcolsep}{4.5pt}
  \vskip 0.1in
  \tiny{
    \begin{tabular}{ccc|ccccccc}
    \toprule
    Task & Split & Languages & \texttt{{jc-v1}} & \texttt{{nllb-b}} & \texttt{{nllb-l}} & \texttt{{je-v3}} & \texttt{{jc-v2-s1}} & \texttt{{jc-v2-s2}} & \texttt{{jc-v2}} \\
    \midrule
    MLQARetrieval & test & hi,de & 8.05 & 21.71 & 39.25 & \textbf{73.69} & 67.16 & 67.60 & 72.32 \\
    MLQARetrieval & test & hi,en & 10.28 & 36.62 & 41.89 & \textbf{65.76} & 56.77 & 58.33 & 64.18 \\
    MLQARetrieval & test & hi,es & 8.33 & 32.66 & 45.79 & \textbf{70.49} & 64.31 & 65.37 & 69.20 \\
    MLQARetrieval & test & hi,zh & 2.62 & 34.75 & 42.37 & \textbf{67.60} & 61.29 & 62.03 & 65.39 \\
    MLQARetrieval & test & zh,de & 14.69 & 18.84 & 33.67 & \textbf{67.35} & 59.91 & 63.49 & 66.37 \\
    MLQARetrieval & test & zh,en & 18.78 & 29.21 & 36.59 & \textbf{60.33} & 48.72 & 53.52 & 59.50 \\
    MLQARetrieval & test & zh,es & 14.23 & 26.81 & 39.87 & \textbf{64.46} & 57.42 & 59.66 & 63.39 \\
    MLQARetrieval & test & zh,hi & 2.43 & 36.52 & 44.44 & \textbf{61.93} & 54.28 & 57.14 & 60.20 \\
    XPQARetrieval & test & en,de & 13.45 & 9.98 & 28.71 & \textbf{60.48} & 58.75 & 60.24 & 60.28 \\
    XPQARetrieval & test & de,en & 39.28 & 29.44 & 47.12 & 82.12 & 82.80 & \textbf{82.95} & 79.45 \\
    XPQARetrieval & test & en,es & 20.41 & 14.32 & 30.48 & \textbf{53.46} & 49.64 & 52.31 & 53.17 \\
    XPQARetrieval & test & es,en & 40.61 & 31.30 & 45.34 & 68.31 & 69.14 & \textbf{70.26} & 67.62 \\
    XPQARetrieval & test & en,fr & 22.20 & 15.66 & 31.46 & 54.80 & 52.15 & \textbf{56.24} & 55.54 \\
    XPQARetrieval & test & fr,en & 47.06 & 36.14 & 49.88 & 73.83 & 75.44 & \textbf{76.56} & 72.96 \\
    XPQARetrieval & test & en,hi & 4.87 & 27.87 & 34.86 & \textbf{41.10} & 36.15 & 38.23 & 39.63 \\
    XPQARetrieval & test & hi,en & 8.07 & 56.85 & 62.61 & \textbf{75.41} & 74.52 & 74.41 & 71.56 \\
    XPQARetrieval & test & en,jp & 5.12 & 17.09 & 28.58 & 47.12 & 47.19 & \textbf{49.49} & 46.51 \\
    XPQARetrieval & test & jp,en & 19.43 & 46.09 & 54.54 & 73.50 & 71.75 & \textbf{73.87} & 69.75 \\
    XPQARetrieval & test & en,pl & 12.08 & 9.86 & 16.86 & 34.19 & 34.23 & \textbf{35.93} & 34.56 \\
    XPQARetrieval & test & pl,en & 20.69 & 17.50 & 25.65 & 50.88 & 51.56 & \textbf{52.74} & 49.98 \\
    XPQARetrieval & test & en,zh & 4.41 & 12.86 & 19.12 & \textbf{37.43} & 34.62 & 36.07 & 36.21 \\
    XPQARetrieval & test & zh,en & 12.96 & 32.39 & 39.80 & \textbf{64.52} & 62.18 & 60.96 & 58.95 \\
    \makecell{mFollowIRCrossLingual \\ InstructionRetrieval} & test & en,ru & 1.04 & \textbf{1.55} & 1.24 & 0.22 & 0.83 & -0.20 & -2.13 \\
    \makecell{mFollowIRCrossLingual \\ InstructionRetrieval} & test & en,zh & 0.18 & -1.27 & -1.47 & -0.69 & \textbf{3.07} & 2.71 & -0.72 \\
    \midrule
    \multicolumn{3}{l|}{\textbf{Average}} & 23.43 & 36.24 & 47.43 & \textbf{69.84} & 66.67 & 68.03 & 68.78 \\
    \midrule
    \midrule
    \multicolumn{10}{c}{\textbf{STS} Spearman correlation on cosine similarity} \\
    \midrule
    IndicCrosslingualSTS & test & en,hi & -11.26 & 58.03 & 58.73 & 67.84 & 62.49 & 61.18 & \textbf{72.51} \\
    STS17 & test & en,de & 53.67 & 54.88 & 65.87 & 83.66 & 84.18 & 84.03 & \textbf{85.89} \\
    STS17 & test & es,en & 52.80 & 64.91 & 77.17 & 85.09 & 85.60 & 85.27 & \textbf{87.28} \\
    STS17 & test & fr,en & 57.46 & 66.04 & 76.21 & 83.77 & 83.51 & 83.47 & \textbf{85.91} \\
    STS22.v2 & test & de,en & 55.96 & 56.41 & 54.70 & 61.63 & 65.06 & \textbf{65.86} & 61.14 \\
    STS22.v2 & test & es,en & 69.92 & 56.79 & 65.06 & 81.80 & 81.14 & \textbf{82.53} & 81.48 \\
    STS22.v2 & test & pl,en & 62.61 & 58.90 & 67.59 & 76.82 & 77.60 & \textbf{82.35} & 78.00 \\
    STS22.v2 & test & zh,en & 39.82 & 57.88 & 60.61 & \textbf{76.33} & 73.93 & 74.16 & 70.47 \\
    STS22.v2 & test & de,fr & 39.37 & 42.33 & 44.69 & 55.67 & 53.96 & \textbf{60.86} & 60.39 \\
    STS22.v2 & test & de,pl & 25.87 & 17.52 & 27.62 & \textbf{56.69} & 39.79 & 42.93 & 47.27 \\
    STS22.v2 & test & fr,pl & \textbf{84.52} & 50.71 & 39.44 & 84.52 & 50.71 & 61.98 & 61.98 \\
    \midrule
    \multicolumn{3}{l|}{\textbf{Average}} & 48.25 & 53.13 & 57.97 & \textbf{73.98} & 68.91 & 71.33 & 72.03 \\
    \midrule
    \bottomrule
    \end{tabular}
    \\
    \vspace{0.1in}
    \footnotesize{Evaluated on all tasks included in MTEB version $\mathrm{1.34.7}$, except the following: \texttt{MrTidyRetrieval}, \texttt{BrightRetrieval}, \texttt{MSMARCOv2}, \texttt{NeuCLIR2022Retrieval}, \texttt{NeuCLIR2023Retrieval} and \texttt{MIRACLRetrieval}. These tasks were excluded either due to bugs in the evaluation code or excessive computation times.} \\
  }
\end{table*}

\begin{table*}[htb]
    \centering
    \setlength{\tabcolsep}{4.5pt} 
    \caption{MRL \citep{mrl} ablation study on the CLIP Benchmark Retrieval tasks}
    \label{appendix:mrl-clipb}
    \vspace{0.1in}
    \begin{center}
    \begin{small}
    \begin{tabular}{l|cccccc}
    \toprule
   Dataset - Dimension & \makecell{1024} & \makecell{768} & \makecell{512} & \makecell{ 256} & \makecell{128} & \makecell{64} \\
    \midrule
    \multicolumn{7}{c}{\textbf{Zero-shot Image Retrieval - Recall@5  [\%]}} \\
    \midrule
    Flickr30K \citep{flickr30k} & 89.84 & \textbf{89.98} & 89.70 & 89.20 & 87.22 & 81.60 \\
    MS COCO \citep{mscococaptions} & \textbf{68.35} & 68.26 & 68.16 & 67.43 & 64.58 & 59.41 \\
    \midrule
    \multicolumn{7}{c}{\textbf{Zero-shot Text Retrieval - Recall@5  [\%]}} \\
    \midrule
    Flickr30K \citep{flickr30k} & 98.00 & \textbf{98.10} & 98.00 & 98.00 & 96.30 & 93.40 \\
    MS COCO \citep{mscococaptions} & \textbf{81.46} & 81.10 & 81.10 & 80.70 & 78.66 & 73.00 \\
    \bottomrule
\end{tabular}
\end{small}
\end{center}
\end{table*}

\begin{table*}[htb]
    \centering
    \setlength{\tabcolsep}{4.5pt} 
    \caption{MRL \citep{mrl} ablation study on Crossmodal-3600 \citep{crossmodal3600}}
    \label{appendix:mrl-crossmodal3600}
    \vspace{0.1in}
    \begin{center}
    \begin{small}
    \begin{tabular}{l|cccccc}
        \toprule
        Language - Dimension & \makecell{1024} & \makecell{768} & \makecell{512} & \makecell{ 256} & \makecell{128} & \makecell{64} \\
        \midrule
        \multicolumn{7}{c}{\textbf{Zero-shot Image Retrieval - Recall@5  [\%]}} \\
        \midrule
        \textbf{Average} & 81.43 & \textbf{82.35} & 82.31 & 81.75 & 78.17 & 72.52 \\
        ar & \textbf{73.56} & 73.39 & 73.17 & 72.61 & 68.42 & 62.28 \\
        bn & \textbf{63.78} & 63.67 & 63.64 & 62.39 & 57.58 & 49.58 \\
        da & \textbf{85.39} & 85.31 & 84.67 & 84.53 & 81.69 & 75.69 \\
        de & 91.25 & 91.28 & \textbf{91.47} & \textbf{91.47} & 88.75 & 84.44 \\
        el & 75.03 & 75.08 & \textbf{75.25} & 74.69 & 72.00 & 66.81 \\
        en & 75.83 & 75.97 & \textbf{76.03} & 75.72 & 74.03 & 69.67 \\
        es & 83.64 & 83.67 & \textbf{83.86} & 83.14 & 81.00 & 76.19 \\
        fi & 82.83 & \textbf{83.03} & 82.61 & 81.67 & 79.44 & 74.94 \\
        fr & \textbf{88.78} & 88.75 & 88.53 & 88.28 & 86.92 & 82.14 \\
        hi & \textbf{55.25} & 54.97 & 55.14 & 54.14 & 50.06 & 42.33 \\
        id & \textbf{84.22} & 84.00 & 84.11 & 83.25 & 80.00 & 74.17 \\
        it & 88.33 & \textbf{88.53} & 88.31 & 87.83 & 85.11 & 79.75 \\
        ja & \textbf{87.03} & \textbf{87.03} & 86.89 & 86.33 & 82.92 & 75.39 \\
        ko & \textbf{78.81} & 78.72 & 78.56 & 77.03 & 73.44 & 65.78 \\
        nl & \textbf{82.56} & \textbf{82.56} & 82.31 & 82.25 & 79.00 & 73.25 \\
        no & \textbf{81.08} & 80.94 & 80.64 & 80.14 & 76.94 & 71.56 \\
        pl & 84.00 & 83.97 & \textbf{84.06} & 83.39 & 81.28 & 76.83 \\
        pt & 82.42 & \textbf{82.50} & 82.00 & 81.58 & 79.36 & 72.97 \\
        ro & 89.36 & \textbf{89.39} & 89.22 & 89.14 & 87.03 & 82.06 \\
        ru & 88.97 & \textbf{89.17} & 89.08 & 88.69 & 87.00 & 82.17 \\
        sv & \textbf{78.06} & 77.86 & 78.00 & 77.47 & 74.83 & 69.53 \\
        th & 81.61 & \textbf{81.64} & 81.14 & 80.36 & 76.97 & 68.17 \\
        tr & 81.31 & \textbf{81.44} & 81.22 & 80.69 & 77.39 & 71.33 \\
        uk & 88.56 & 88.42 & \textbf{88.61} & 87.89 & 85.86 & 80.72 \\
        vi & 86.64 & \textbf{86.72} & 86.69 & 85.92 & 83.17 & 77.86 \\
        zh & 78.97 & \textbf{79.06} & 78.86 & 78.08 & 74.81 & 67.36 \\
        \midrule
        \multicolumn{7}{c}{\textbf{Zero-shot Text Retrieval - Recall@5  [\%]}} \\
        \midrule
        \textbf{Average} & 83.23 & \textbf{83.26} & 83.21 & 82.81 & 80.54 & 75.37 \\
        ar & \textbf{76.25} & 76.17 & 76.17 & 75.19 & 73.61 & 68.25 \\
        bn & \textbf{69.00} & \textbf{69.00} & 68.94 & 68.75 & 66.19 & 59.39 \\
        da & \textbf{88.53} & 88.39 & 88.25 & 87.58 & 85.69 & 80.25 \\
        de & 92.47 & \textbf{92.56} & 92.42 & 92.31 & 90.64 & 87.19 \\
        el & 73.33 & \textbf{73.50} & 73.47 & 73.39 & 72.75 & 68.61 \\
        en & 78.58 & \textbf{78.61} & 78.36 & 77.86 & 76.44 & 72.58 \\
        es & 86.28 & \textbf{86.39} & 86.33 & 85.78 & 84.03 & 80.17 \\
        fi & 84.19 & \textbf{84.25} & 84.00 & 83.58 & 82.08 & 77.81 \\
        fr & \textbf{90.89} & 90.86 & 90.75 & 90.14 & 88.61 & 84.89 \\
        hi & \textbf{61.64} & \textbf{61.64} & 61.44 & 61.44 & 58.86 & 52.81 \\
        id & \textbf{86.31} & 86.25 & 86.19 & 86.14 & 83.92 & 80.19 \\
        it & \textbf{90.17} & 90.08 & \textbf{90.17} & 89.83 & 87.78 & 83.75 \\
        ja & 88.50 & \textbf{88.64} & \textbf{88.64} & 87.92 & 86.22 & 81.64 \\
        ko & \textbf{81.42} & \textbf{81.42} & 81.31 & 80.53 & 78.31 & 73.03 \\
        nl & \textbf{82.47} & 82.44 & 82.36 & 81.94 & 79.97 & 76.00 \\
        no & 83.75 & \textbf{83.81} & 83.78 & 83.28 & 80.39 & 75.39 \\
        pl & 84.61 & \textbf{84.67} & 84.47 & 83.81 & 82.36 & 78.56 \\
        pt & \textbf{83.94} & 83.83 & 83.64 & 83.25 & 81.72 & 77.00 \\
        ro & \textbf{91.31} & 91.14 & 91.22 & 91.03 & 89.44 & 85.67 \\
        ru & \textbf{90.64} & 90.58 & 90.31 & 90.19 & 88.50 & 84.86 \\
        sv & \textbf{78.28} & 78.22 & 78.06 & 77.50 & 75.97 & 71.42 \\
        th & \textbf{81.94} & 81.89 & \textbf{81.94} & 81.28 & 79.25 & 73.61 \\
        tr & 82.67 & \textbf{82.72} & 82.39 & 81.92 & 80.36 & 76.06 \\
        uk & \textbf{89.53} & \textbf{89.53} & 89.31 & 88.81 & 88.00 & 84.69 \\
        vi & \textbf{88.06} & 87.94 & 87.61 & 87.47 & 85.92 & 82.00 \\
        zh & \textbf{79.22} & 79.06 & 78.83 & 78.50 & 76.11 & 70.22 \\
        \bottomrule
\end{tabular}
\end{small}
\end{center}
\end{table*}

\begin{table*}[htb]
    \centering
    \setlength{\tabcolsep}{4.5pt} 
    \caption{MRL \citep{mrl} ablation study on XTD10 \citep{xtd10, mic}}
    \label{appendix:mrl-xtd10}
    \vspace{0.1in}
    \begin{center}
    \begin{small}
    \begin{tabular}{l|cccccc}
    \toprule
    Language - Dimension & \makecell{1024} & \makecell{768} & \makecell{512} & \makecell{ 256} & \makecell{128} & \makecell{64} \\
    \midrule
    \multicolumn{7}{c}{\textbf{Zero-shot Image Retrieval - Recall@5  [\%]}} \\
        \midrule
        \textbf{Average} & \textbf{84.87} & 84.85 & 84.60 & 84.32 & 81.80 & 77.85 \\
        de & \textbf{85.70} & 85.40 & 84.90 & 84.70 & 81.10 & 79.30 \\
        en & 89.40 & \textbf{89.60} & 88.70 & 88.90 & 86.50 & 83.00 \\
        es & 85.90 & 85.90 & \textbf{86.00} & 85.70 & 84.00 & 80.80 \\
        fr & 85.10 & 84.70 & 84.80 & \textbf{85.30} & 83.30 & 79.10 \\
        it & 85.80 & 85.50 & \textbf{86.10} & 85.80 & 83.70 & 81.10 \\
        ko & \textbf{82.10} & 81.90 & \textbf{82.10} & 80.10 & 77.90 & 71.90 \\
        pl & 86.50 & \textbf{86.70} & 86.40 & 86.30 & 84.80 & 80.40 \\
        ru & 81.10 & 81.20 & 80.80 & \textbf{81.40} & 77.80 & 74.00 \\
        tr & 83.70 & \textbf{84.10} & 83.50 & 83.10 & 80.00 & 75.80 \\
        zh & 83.40 & \textbf{83.50} & 82.70 & 81.90 & 78.90 & 73.10 \\
        \midrule
        \multicolumn{7}{c}{\textbf{Zero-shot Text Retrieval - Recall@5  [\%]}} \\
        \midrule
        \textbf{Average} & \textbf{86.03} & 86.02 & 86.02 & 85.84 & 84.37 & 81.02 \\
        de & 86.20 & \textbf{86.40} & 86.10 & 85.90 & 85.20 & 81.50 \\
        en & \textbf{90.50} & \textbf{90.50} & 91.10 & 90.40 & 90.10 & 86.80 \\
        es & \textbf{87.00} & 86.80 & 86.70 & 86.90 & 86.10 & 81.80 \\
        fr & \textbf{85.20} & 84.90 & \textbf{85.20} & 85.00 & 84.20 & 81.70 \\
        it & \textbf{88.00} & \textbf{88.00} & 87.70 & 87.60 & 85.70 & 83.20 \\
        ko & 82.10 & 82.20 & 82.40 & \textbf{82.80} & 80.30 & 77.50 \\
        pl & 88.50 & \textbf{88.70} & 88.60 & 88.30 & 85.90 & 83.60 \\
        ru & \textbf{83.10} & 83.00 & 82.70 & 82.80 & 80.80 & 78.10 \\
        tr & 85.60 & 85.60 & \textbf{85.90} & \textbf{85.90} & 84.50 & 79.80 \\
        zh & \textbf{84.10} & \textbf{84.10} & 83.80 & 82.80 & 80.90 & 76.20 \\
        \bottomrule
\end{tabular}
\end{small}
\end{center}
\end{table*}

\begin{table*}[htb]
    \centering
    \setlength{\tabcolsep}{4.5pt} 
    \caption{MRL \citep{mrl} ablation study on English Classic MTEB \citep{mteb} Retrieval and STS tasks}
    \vspace{0.1in}
    \label{appendix:mrl-mteb}
    \begin{center}
    \begin{small}
    \begin{tabular}{l|cccccc}
    \toprule
    Dataset - Dimensions & \makecell{1024} & \makecell{768} & \makecell{512} & \makecell{256} & \makecell{128} & \makecell{64}  \\
    \midrule
    \multicolumn{7}{c}{\textbf{Retrieval - nDCG@10}} \\
    \midrule
    \textbf{Average} & \textbf{49.33} & 49.32 & 49.19 & 48.67 & 46.37 & 40.66 \\
    FiQA2018         & 41.93 & \textbf{42.04} & 41.79 & 40.96 & 38.75 & 34.27 \\
    NFCorpus         & 32.89 & \textbf{33.01} & 32.72 & 32.32 & 30.20 & 25.69 \\
    SciFact          & \textbf{65.34} & 65.12 & 65.26 & 64.57 & 62.51 & 57.17 \\
    SCIDOCS          & \textbf{18.90} & 18.85 & 18.84 & 18.46 & 16.98 & 13.84 \\
    CQADupstackRetrieval & \textbf{41.20} & 41.17 & 41.06 & 40.47 & 37.97 & 32.42 \\
    Touche2020       & 23.94 & 24.08 & 24.23 & 24.57 & \textbf{25.82} & 23.26 \\
    TRECCOVID        & \textbf{76.73} & 76.54 & 75.98 & 76.01 & 73.10 & 67.47 \\
    FEVER            & 84.96 & \textbf{84.99} & 84.83 & 84.37 & 81.40 & 69.26 \\
    HotpotQA         & \textbf{60.14} & 60.13 & 59.71 & 58.06 & 52.14 & 38.97 \\
    DBPedia          & \textbf{37.44} & 37.42 & 36.98 & 36.12 & 33.21 & 26.34 \\
    NQ               & 57.17 & \textbf{57.19} & 57.02 & 56.30 & 53.49 & 45.42 \\
    ClimateFEVER     & 30.12 & 30.15 & \textbf{30.35} & 30.00 & 26.08 & 20.89 \\
    MSMARCO          & \textbf{37.47} & 37.43 & 37.45 & 37.16 & 35.74 & 32.47 \\
    ArguAna          & \textbf{43.55} & 43.50 & 43.53 & 42.83 & 41.06 & 37.21 \\
    QuoraRetrieval   & \textbf{88.14} & 88.13 & 88.06 & 87.88 & 87.10 & 85.19 \\
    \midrule
    \multicolumn{7}{c}{\textbf{STS - Spearman correlation based on cosine similarity}} \\
    \midrule
    \textbf{Average}      & \textbf{81.29} & 81.27 & 81.26 & 81.24 & 80.78 & 79.56 \\
    STS12        & 76.72 & 76.72 & 76.85 & 76.97 & \textbf{77.13} & 76.44 \\
    STS13        & 79.90 & 79.89 & 79.98 & \textbf{80.48} & 80.00 & 78.69 \\
    STS14        & 77.49 & 77.51 & 77.57 & \textbf{77.67} & 77.10 & 75.50 \\
    STS15        & 86.42 & 86.43 & 86.46 & \textbf{86.58} & 86.21 & 84.58 \\
    STS16        & \textbf{85.18} & 85.17 & 85.15 & 85.10 & 84.52 & 84.03 \\
    STS17        & 87.87 & \textbf{87.89} & 87.74 & 87.31 & 86.91 & 85.36 \\
    STS22        & \textbf{67.07} & 67.04 & 67.01 & 66.95 & 66.66 & 66.71 \\
    BIOSSES      & 83.00 & \textbf{83.03} & 82.65 & 82.33 & 81.18 & 78.27 \\
    STSBenchmark & \textbf{86.87} & 86.68 & 86.86 & 86.72 & 86.26 & 85.14 \\
    SICK-R       & \textbf{82.38} & \textbf{82.38} & 82.36 & 82.27 & 81.87 & 80.89 \\
    \bottomrule
    \end{tabular}
    \end{small}
    \end{center}
\end{table*}

\begin{figure}[h]
    \centering
    \includegraphics[width=0.9\textwidth]{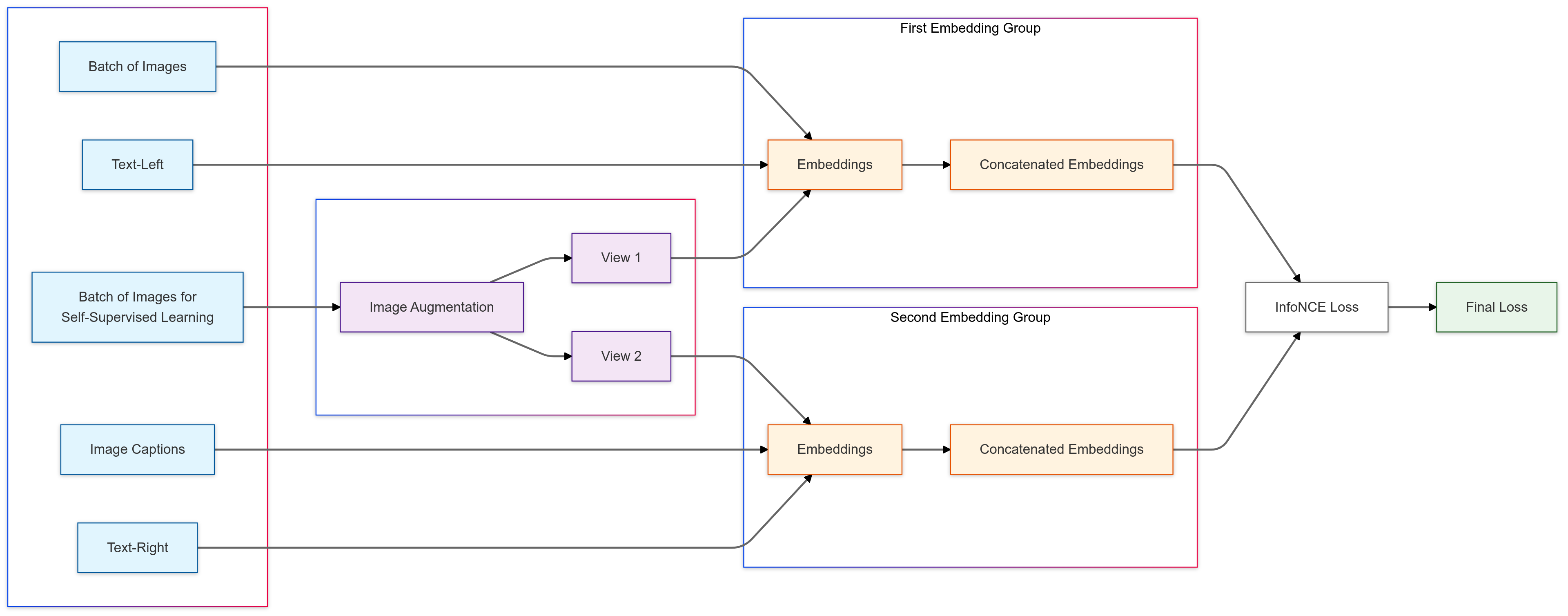}
    \caption{Contrastive learning between two embedding groups (Unified Batch technique). The first group concatenates original images, question texts, and one view of augmented images, while the second group concatenates corresponding image captions, answer texts, and a different view of the augmented images.}
    \label{appendix:unified-batch}
\end{figure}

\end{document}

%% file: math_commands.tex
\usepackage{amsmath,amsfonts,bm}

\def\eqref#1{equation~\ref{#1}}
\def\1{\bm{1}}

\DeclareMathAlphabet{\mathsfit}{\encodingdefault}{\sfdefault}{m}{sl}
\SetMathAlphabet{\mathsfit}{bold}{\encodingdefault}{\sfdefault}{bx}{n}